\documentclass{article}

\usepackage[preprint,eandd]{neurips_2026}

\usepackage[utf8]{inputenc}
\usepackage[T1]{fontenc}
\usepackage{hyperref}
\usepackage{url}
\usepackage{doi}
\usepackage{booktabs}
\usepackage{amsmath}
\usepackage{amsfonts}
\usepackage{graphicx}
\usepackage{multirow}
\usepackage{pifont}
\usepackage{xcolor}
\usepackage{newunicodechar}
\newunicodechar{，}{,}
\usepackage{microtype}
\usepackage{tikz}
\usepackage{float}
\usetikzlibrary{arrows.meta}
\usepackage{enumitem}
\setcitestyle{numbers,square,comma}
\newcommand{\cmark}{\ding{51}}
\newcommand{\xmark}{\ding{55}}
\newcommand{\pmark}{$\triangle$}

\usepackage{authblk}
\title{PhysEditWorld: A Large-Scale Dataset Toward Physics-Editable World Models}

\author[1,*]{Bin Hu}
\author[2,*]{Yanwen Ma}
\author[3,$\ddagger$]{Jiehui Huang}
\author[4]{Ziliang Zhang}
\author[5]{Haoning Wu}
\author[1]{Ruicheng Zhang}
\author[6]{Yaokun Li}
\author[6]{Zijun Wang}
\author[7]{Yuechen Zhang}
\author[1]{Chun-Mei Tseng}
\author[6]{Hanhui Li}
\author[8]{Shengju Qian}
\author[1]{Jun Zhou}
\author[9]{Kaipeng Zhang}
\author[6]{Xiaodan Liang}
\author[3,$\dagger$]{Jiaya Jia}
\author[1,$\dagger$]{Xiu Li}

\affil[1]{Tsinghua Shenzhen International Graduate School, Tsinghua University}
\affil[2]{Beihang University}
\affil[3]{The Hong Kong University of Science and Technology}
\affil[4]{Independent Researcher}
\affil[5]{Shanghai Jiao Tong University}
\affil[6]{Sun Yat-sen University}
\affil[7]{The Chinese University of Hong Kong}
\affil[8]{Alaya Studio}
\affil[9]{Tencent}

\affil[*]{\protect\vspace{1.5em}Equal contribution \quad \textsuperscript{$\ddagger$}Project lead \quad \textsuperscript{\dag}Corresponding authors}

\begin{document}
\maketitle

\begin{abstract}
Recent game world models can synthesize visually plausible, action-conditioned rollouts. However, their interaction behaviors often remain limited to exploratory or wandering trajectories, and physical dynamics are typically learned as implicit correlations from data rather than as controllable variables. This limitation hinders their applicability to authored game environments, where physical rules are deliberately designed and require explicit manipulation. We introduce PhysEditWorld, a multimodal dataset with physical parameters, with a primary focus on gravity in this initial version. At its core, PhysEditWorld is built upon a  replay paradigm implemented with a UE5 replay-and-rendering pipeline. Each scenario records a normalized action trace and replays the same initial state, character controller, action sequence, and camera policy under multiple gravity configurations, enabling controlled and attributable physical variation. PhysEditWorld contains 12 cinematic UE5 scenes, over 100 hours of gameplay interactions, and more than 60 million rendered rollout frames. Each sample provides synchronized multimodal signals, including RGB, depth, normals, audio, action traces, camera trajectory, engine states, semantic annotations, and explicit gravity labels. We further conduct initial utility studies on both generative video models and world understanding models, demonstrating that PhysEditWorld enables improved gravity-faithful dynamics modeling, enhances  consistency under physical edits, and provides a scalable foundation for controllable world modeling research. \\
\textbf{Project page:} \url{https://yizhiqianbi.github.io/physeditworld/}.
\end{abstract}

\begin{figure}[htbp]
    \centering
    \includegraphics[width=0.96\linewidth]{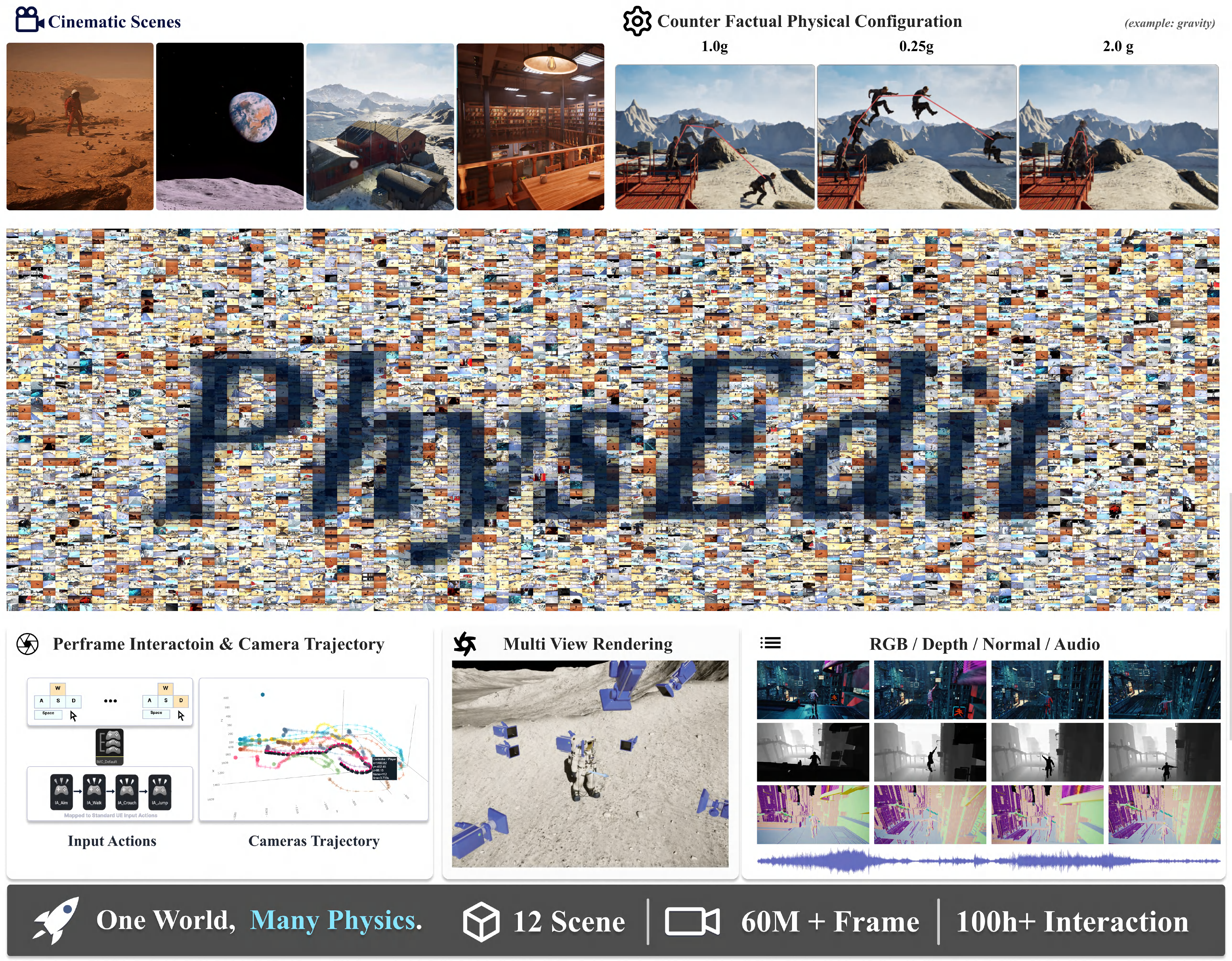}
    \caption{Overview of PhysEditWorld. The dataset contains 12 cinematic UE5 scenes, 100+ hours of gameplay interactions, and 60M+ rendered rollout frames. Recorded interactions are replayed under editable physical configurations and rendered from 8 synchronized camera views, with RGB, depth, normal, audio, action, camera, engine-state, and gravity annotations.}
    \label{fig:teaser}
\end{figure}

\section{Introduction}

Recent game world models have progressed from visual predictors to interactive generative simulators. Systems such as Genie, DIAMOND, GameNGen, GameGen-X, YUMI, LingBot-World, and Matrix-Game-3.0 show that large generative models can synthesize plausible gameplay trajectories, support user interaction, or maintain longer-horizon world consistency \citep{bruce2024genie,alonso2024diamond,valevski2024gamengen,che2024gamegenx,lingbotworld2026,matrixgame2026,mao2025yume}. However, these models typically learn physics as an implicit regularity of the data distribution. They can imitate how a game usually evolves, but they are not designed to answer a question that is central to game authoring: how should the same scene evolve if a physical rule is edited?

Unlike natural world modeling, where physics is usually treated as a latent constant, game world modeling must eventually support physical laws as editable design variables. Developers routinely tune gravity scale, jump behavior, friction, drag, and wind to shape pacing, difficulty, player feel, and emergent motion. A learned world model that entangles these parameters with appearance and gameplay statistics may generate visually plausible clips, yet still fail as an editable game simulator. In this paper, we focus on gravity as a first measurable step toward editable physics, because it is widely supported across engines and produces observable changes in jump arcs, airtime, fall speed, and object trajectories.

Existing datasets and benchmarks do not directly measure this capability. Game and reinforcement-learning environments such as ALE, Procgen, MineRL, and CARLA provide interactive worlds or human demonstrations, but they are primarily designed for policy learning, navigation, or generalization under fixed rules \citep{bellemare2013ale,cobbe2020procgen,guss2019minerl,dosovitskiy2017carla}. World-exploration datasets such as Sekai provide large-scale first-person and drone-view videos with rich annotations and camera trajectories, but are not organized around matched physical interventions \citep{li2025sekai}. Physics reasoning and video-generation benchmarks such as PHYRE, CLEVRER, Physion, VBench, VideoPhy, PhyGenBench, Physics-IQ, and PhysInOne evaluate intuitive physics, physical plausibility, or physics-aware generation \citep{bakhtin2019phyre,yi2020clevrer,bear2021physion,huang2024vbench,bansal2025videophy,meng2025phygenbench,motamed2025physicsiq,zhou2026physinone}. These resources are valuable, but they do not provide matched gameplay clips in which the scene, interaction trace, and camera policy are held fixed while gravity is explicitly changed.

We introduce \textbf{PhysEditWorld}, a multimodal dataset for physics-editable game world modeling. PhysEditWorld contains 12 cinematic UE5 scenes and over 60M rendered frames, with matched rollouts designed for gravity-conditioned generation and evaluation. The dataset is built around \emph{comparability}: each replay group fixes the authored scene, initial state, action trace, character controller, and camera policy, then reruns the scenario under different gravity configurations. The pipeline records synchronized first-person RGB, third-person RGB, depth, normal maps, action traces, semantic captions, engine states, and gravity annotations. Because non-gravity factors are controlled within a replay group, the resulting motion differences can be evaluated as responses to the physical intervention.

PhysEditWorld enables evaluation beyond visual plausibility by testing whether generated rollouts are faithful to edited gravity. We study this capability in gravity-conditioned generation, first-person world-model rollouts, and video-language gravity inference. Across representative backbones, we find that current models can maintain visual realism, but often under-express gravity-sensitive motion or confuse the relative ordering of gravity levels. This suggests that editable physics remains a missing capability in current game world models.

\paragraph{Our contributions are as follows.}
\begin{itemize}[leftmargin=1.2em]
    \item We introduce \textbf{PhysEditWorld}, a large-scale multimodal dataset for physics-editable game world modeling. To the best of our knowledge, it is the first dataset organized around matched gameplay rollouts with explicit gravity interventions.
    \item We develop a UE5 \textbf{replay-and-rendering pipeline} that automatically replays the same scene, initial state, character controller, action trace, and camera policy under controlled physics configurations.
    \item We conduct dataset utility studies showing that PhysEditWorld can be used to evaluate and improve gravity awareness in generative video models, game world models, and video-language models.
\end{itemize}

\section{Related work}

\textbf{Game world models.}
Game world models have evolved from single-game neural simulators to open-domain interactive video generators. Early works such as GameGAN, Playable Video Generation, Playable Environments, and Promptable Game Models studied controllable neural simulation and interactive video manipulation from gameplay or video data~\citep{kim2020gamegan,menapace2021pvg,menapace2022playableenv,menapace2024pgm,xu2026eponav2}. Recent systems, including Genie, Oasis, DIAMOND, GameNGen, GameGen-X, GameFactory, MineWorld, YUME, LingBot-World, Matrix-Game 3.0, and related interactive video world models, scale action-conditioned generation, open-world exploration, long-horizon consistency, ID consistency, and real-time interaction~\citep{bruce2024genie,quevedo2024oasis,alonso2024diamond,valevski2024gamengen,che2024gamegenx,yu2025gamefactory,guo2025mineworld,mao2025yume,lingbotworld2026,matrixgame2026,chen2025interactiveworldmodels,hu2026identityconsistentvideogenerationlarge}. While these models learn rich visual and interaction dynamics, physical rules are usually absorbed as implicit dataset regularities rather than exposed as editable variables. PhysEditWorld addresses this gap with explicit gravity annotations and matched replays under fixed scene, action, controller, and camera conditions.

\textbf{Video and world-model datasets.}
Existing game and world-model datasets mainly emphasize scale, action control, exploration coverage, or state supervision. MineRL, VPT, and MineDojo provide Minecraft demonstrations, video pretraining data, simulator environments, or task suites for embodied agent learning~\citep{guss2019minerl,baker2022vpt,fan2022minedojo}. OGameData/GameGen-X and GF-Minecraft/GameFactory target open-world or action-controllable game video generation~\citep{che2024gamegenx,yu2025gamefactory}; Sekai and YUME focus on world exploration and interactive generation~\citep{li2025sekai,mao2025yume}; and WildWorld and MultiWorld add explicit state, multi-agent, or multi-view supervision~\citep{li2026wildworld,wu2026multiworld}. Related resources also study high-DoF action-to-video control or physics-related gameplay failures~\citep{wang2025visualactionprompts,taesiri2022clipgamephysics}. However, these datasets are not organized around matched physical interventions. PhysEditWorld instead replays the same authored scene, interaction trace, character controller, and camera policy under different gravity configurations, making the physical edit directly comparable.

\textbf{Physics benchmarks and controllable generation.}
Many benchmarks study physical understanding from visual data. ShapeStacks and ADEPT probe object stability and expectation violation, while PHYRE, I-PHYRE, IntPhys, IntPhys2, CLEVRER, CATER, and Physion evaluate intuitive physics, compositional actions, causal reasoning, intervention, and future prediction~\citep{groth2018shapestacks,smith2019adept,bakhtin2019phyre,li2024iphyre,riochet2022intphys,bordes2025intphys2,yi2020clevrer,girdhar2020cater,bear2021physion}. CoPhy, ComPhy, CRIPP-VQA, ContPhy, CausalVQA, and QuantiPhy further study counterfactual dynamics, hidden physical properties, causal alternatives, and quantitative physical quantities~\citep{baradel2020cophy,chen2022comphy,patel2022crippvqa,zheng2024contphy,foss2025causalvqa,li2025quantiphy}. Video-generation benchmarks such as VBench, VBench-2.0, VideoPhy, PhyGenBench, Physics-IQ, PhysInOne, WorldScore, and NewtonRewards evaluate physical plausibility, intrinsic faithfulness, world-generation quality, or Newtonian motion consistency~\citep{huang2024vbench,zheng2025vbench2,bansal2025videophy,meng2025phygenbench,motamed2025physicsiq,zhou2026physinone,duan2025worldscore,le2025newtonrewards,zhang2026mindvhierarchicalworldmodel,zhang2026robostereodualtower4dembodied}. Recent controllable generation methods condition on force, torque, force fields, physical parameters, Newtonian dynamics, or learned physical priors~\citep{liu2024physgen,gillman2025forceprompting,wang2025physctrl,zhang2025physchoreo,yuan2026newtongen,narayanan2026phyco,li2025pisa}. These works provide important reasoning, evaluation, and generation tools, but they do not provide interactive game rollouts where one physical rule is edited while scene content and interaction are held fixed.

\textbf{Synthetic simulation platforms.}
Simulation platforms enable controllable data generation and ground-truth annotation. ALE, Procgen, MineRL, and CARLA support standardized environments for games, reinforcement learning, and driving~\citep{bellemare2013ale,cobbe2020procgen,guss2019minerl,dosovitskiy2017carla}. AI2-THOR, RoboTHOR, ProcTHOR, Habitat, Gibson, iGibson, TDW, Kubric, VirtualHome, and BEHAVIOR-1K provide controllable 3D simulation with rich sensor outputs, physical interaction, programmatic actions, or scalable embodied-AI environments~\citep{kolve2017ai2thor,deitke2020robothor,deitke2022procthor,savva2019habitat,xia2018gibson,li2022igibson2,gan2021tdw,greff2022kubric,puig2018virtualhome,huang2025dc,li2024behavior1k}. Robotics simulators such as SAPIEN, RLBench, ManiSkill2, and Isaac Gym support articulated-object interaction, manipulation, and high-throughput physics simulation~\citep{xiang2020sapien,james2020rlbench,gu2023maniskill2,makoviychuk2021isaacgym}. UnrealCV connects Unreal Engine to computer-vision pipelines, while UnrealZoo scales UE environments for embodied AI~\citep{qiu2016unrealcv,zhong2025unrealzoo}. PhysEditWorld builds on UE but targets a different protocol: matched replay under edited gravity.

\begin{table}[H]
  \centering
  \caption{Comparison with related datasets and data-generation resources. RGB-D-N-A denotes synchronized RGB, depth, normal, and audio. Edit phys. denotes whether physical attributes are explicitly editable; PhysEditWorld focuses on gravity in the current release. Camera ann. denotes camera parameters or trajectories. Action/control denotes recorded actions, controls, or interaction traces. \cmark: yes, \xmark: no, \pmark: partial.}
  \label{tab:dataset-comparison}
  \resizebox{\linewidth}{!}{%
  \begin{tabular}{lccccc}
    \toprule
    Resource
    & Multi-view
    & RGB-D-N-A
    & Edit phys.
    & Camera ann.
    & Action \\
    \midrule
    ALE \citep{bellemare2013ale}
    & \xmark & \xmark & \xmark & \xmark & \cmark \\
    Procgen \citep{cobbe2020procgen}
    & \xmark & \xmark & \xmark & \xmark & \cmark \\
    MineRL \citep{guss2019minerl}
    & \xmark & \xmark & \xmark & \pmark & \cmark \\
    Sekai \citep{li2025sekai}
    & \pmark & \pmark & \xmark & \cmark & \xmark \\
    GameFactory / GF-Minecraft \citep{yu2025gamefactory}
    & \xmark & \xmark & \xmark & \xmark & \cmark \\
    OGameData / GameGen-X \citep{che2024gamegenx}
    & \pmark & \xmark & \xmark & \cmark & \pmark \\
    \midrule
    CLEVRER \citep{yi2020clevrer}
    & \xmark & \xmark & \xmark & \xmark & \xmark \\
    Physion \citep{bear2021physion}
    & \xmark & \pmark & \xmark & \pmark & \xmark \\
    VBench \citep{huang2024vbench}
    & \xmark & \xmark & \xmark & \xmark & \xmark \\
    PhyGenBench \citep{meng2025phygenbench}
    & \xmark & \xmark & \xmark & \xmark & \xmark \\
    VideoPhy \citep{bansal2025videophy}
    & \xmark & \xmark & \xmark & \xmark & \xmark \\
    Physics-IQ \citep{motamed2025physicsiq}
    & \xmark & \xmark & \xmark & \xmark & \xmark \\
    PhysInOne \citep{zhou2026physinone}
    & \cmark & \pmark & \xmark & \pmark & \xmark \\
    \midrule
    \textbf{PhysEditWorld}
    & \cmark & \cmark & \cmark & \cmark & \cmark \\
    \bottomrule
  \end{tabular}}
\end{table}

\section{ Physics Edit Dataset}
\label{sec:dataset}

PhysEditWorld is a large-scale multimodal dataset for gravity-editable game world modeling. It contains 12 cinematic-quality UE5 scenes, more than 100 hours of human gameplay interactions, and over 60 million rendered rollout frames across multiple gravity configurations and synchronized camera views. All videos are rendered at 30 FPS and $1280 \times 720$ resolution. Unlike many physics-video datasets that focus on short, isolated physical events, PhysEditWorld is built from interactive game scenarios, enabling gravity effects to be studied in action-conditioned world-model rollouts.

Each rollout provides synchronized RGB video, depth maps, surface normals, audio when available, semantic captions, action traces, camera parameters, engine-state logs, and explicit gravity labels. The engine logs include camera trajectories, character states, object states, and relevant physical variables exported in UE5's native world coordinate system and physical scale. The basic unit is a \emph{matched replay group}: the scene, initial state, character controller, action trace, and camera policy are fixed, while only the gravity configuration changes. This structure makes gravity-dependent effects such as jump height, airtime, fall speed, landing timing, camera displacement, and object motion directly comparable across variants.

Table~\ref{tab:dataset-comparison} summarizes the advantage of PhysEditWorld over prior game, world-model, and physics-oriented datasets. Physics-video and physical-reasoning benchmarks such as CLEVRER, Physion, VBench, VideoPhy, PhyGenBench, Physics-IQ, and PhysInOne provide useful tests of physical plausibility, intuitive physics, or physics-aware generation, but generally lack interactive action traces and matched physical interventions \citep{yi2020clevrer,bear2021physion,huang2024vbench,bansal2025videophy,meng2025phygenbench,motamed2025physicsiq,zhou2026physinone}. Game and world-modeling datasets such as ALE, Procgen, MineRL, Sekai, GameFactory/GF-Minecraft, and OGameData/GameGen-X provide action control, visual scale, or camera information to varying degrees, but typically operate under fixed physical rules \citep{bellemare2013ale,cobbe2020procgen,guss2019minerl,li2025sekai,yu2025gamefactory,che2024gamegenx}. In contrast, PhysEditWorld combines multi-view rendering, RGB-depth-normal-audio supervision, explicit editable gravity, camera annotations, and action/control traces, making it possible to evaluate whether a model changes world dynamics consistently under the same scene, action sequence, and camera policy.

\section{Physics Data Pipeline}
\label{sec:data-pipeline}
PhysEditWorld is generated by a UE5 replay-and-rendering pipeline integrated into the standard game-development workflow. Rather than reconstructing scenes in a separate simulator, the pipeline operates directly on artist-ready UE5 levels through an in-editor plug-in, converting authored game content into data-production-ready scenarios with registered scenes, validated assets, controller bindings, camera policies, and replay settings. It then builds normalized action sequences from human-controlled, agent-generated, or scripted inputs and replays them through the same UE5 gameplay stack under controlled physical configurations. The final stage performs synchronized rendering, runtime logging, data organization, and post-hoc VLM annotation, producing aligned rollouts with semantic descriptions.
\begin{figure}[t]
  \centering
  \includegraphics[width=\linewidth]{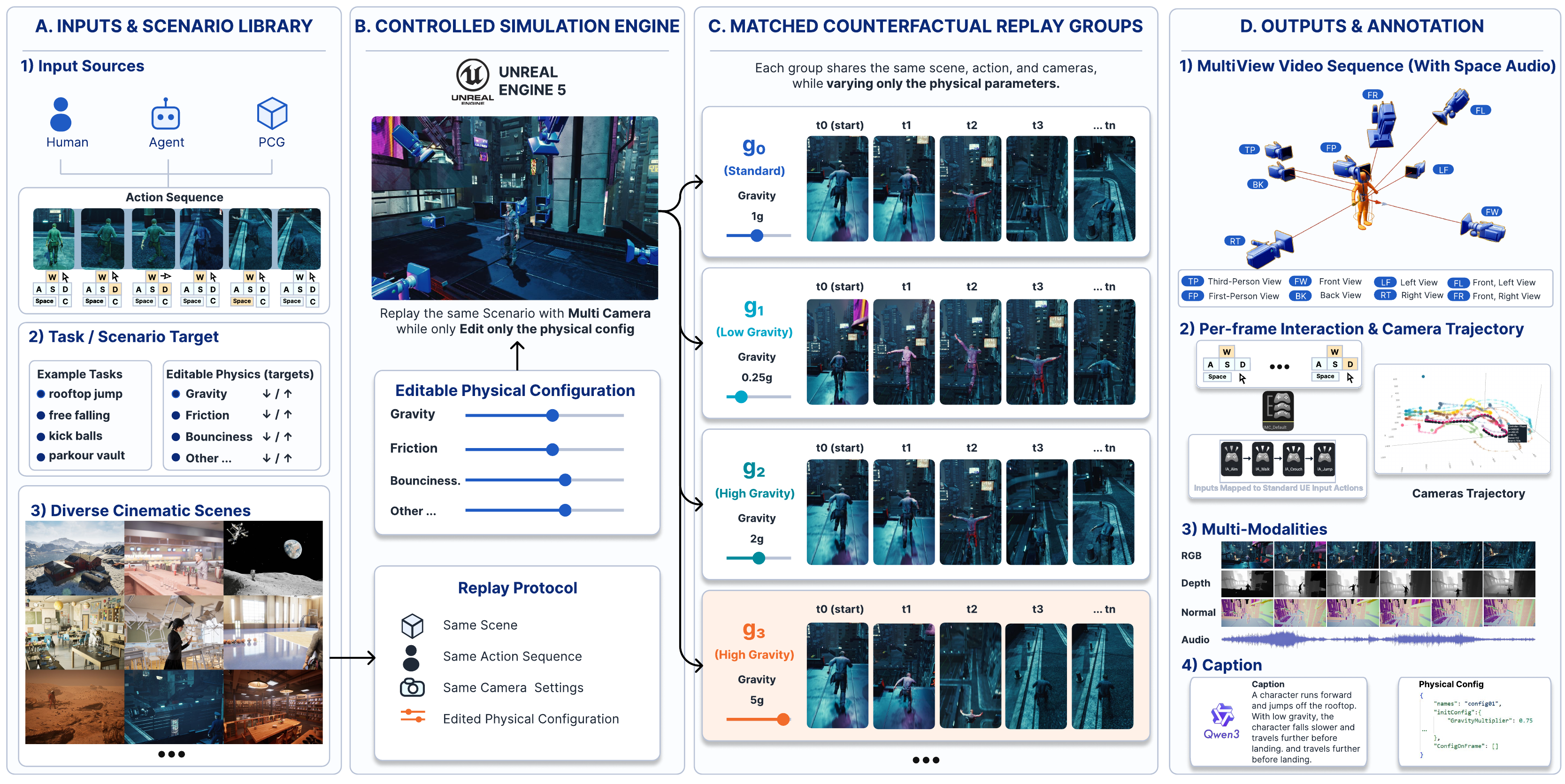}
  \caption{Overview of the PhysEditWorld data pipeline. A) Input sources and scenario library provide human-controlled, agent-generated, or scripted action sequences in diverse cinematic UE5 scenes. B) A controlled UE5 simulation engine replays the same scenario while editing only the physical configuration. C) Matched replay groups share the same scene, action sequence, character controller, and camera policy while varying gravity. D) The pipeline exports synchronized multi-view videos, spatial audio, per-frame interaction and camera trajectories, multimodal render passes, captions, and physical annotations.}
  \label{fig:pipeline}
\end{figure}
\subsection{Scenario and input construction}

The pipeline starts from artist-ready UE5 levels and converts them into replayable scenarios through the in-editor plug-in. Each scenario registers the authored level, interactable assets, character setup, collision configuration, camera anchors, and replay settings required for controlled simulation. Scenario preparation emphasizes physical visibility and replay stability: artists select events such as jumps, falls, object interactions, and height-varying traversal, and validate that the corresponding assets, controllers, spawn points, and camera anchors support reproducible replay.

For each prepared scenario, the pipeline constructs action sequences from human-controlled, agent-generated, or scripted inputs. Instead of recording raw device events, it records semantic Input Action sequences from UE5's Enhanced Input System, including movement axes, jump commands, camera deltas, and key or button states. This representation preserves the intent of the interaction while avoiding hardware-specific confounders such as keyboard layout, mouse sensitivity, controller mapping, or platform-dependent event APIs.

\subsection{Controlled UE5 simulation}

To scale data production within standard game-development workflows, PhysEditWorld uses an in-editor DataFactory plug-in rather than a separate simulator. The plug-in operates on native UE5 abstractions, including authored levels, character controllers, Enhanced Input actions, camera components, collision volumes, and gameplay logic. It provides unified tools for scene registration, replay specification, controller binding, camera setup, physical-parameter editing, validation, and batch execution.

During simulation, the normalized action sequence is injected back into the same UE5 gameplay stack used during capture. Physical configurations are applied as explicit simulation parameters, while the authored scene, controller logic, input sequence, and camera policy remain fixed. This design separates interaction capture from physical intervention and allows controlled physical edits without recollecting behavior or modifying the authored level.

\subsection{Matched replay expansion}

Let $A$ denote the prepared UE5 scenario, $S$ the normalized action sequence, $M$ the character controller, $\theta$ the editable physical configuration, and $\pi_c$ the camera policy. A rollout is generated as
\begin{equation}
  x = F(A,S,M,\theta,\pi_c).
\end{equation}
Replay expansion keeps $(A,S,M,\pi_c)$ fixed and reruns the same scenario under different values of $\theta$. This construction makes the physical edit the controlled variable, so that differences across variants can be attributed to the edited simulation parameter rather than to changes in scene content, input behavior, character setup, or camera specification.

The action sequence is treated as scene-bound because its semantics depend on local geometry, obstacles, and interaction targets. The pipeline therefore expands validated scenario-action pairs over compatible physical configurations and character setups, producing matched variants while preserving the original authored environment and interaction intent.

\subsection{Synchronized export and annotation}
\label{sec:outputs-annotation}

Each replay is exported through Movie Render Queue and runtime logging. Movie Render Queue produces rendered observation streams and auxiliary render passes, while the runtime logger records time-aligned interaction, camera, state, and physical-configuration metadata. All exported records are indexed by a shared rollout identity and frame timeline, enabling deterministic alignment between rendered observations and engine-side logs.

After rendering, the pipeline performs data organization, post-hoc semantic annotation, and quality filtering. We annotate each rollout with a per-clip caption generated by Qwen3-VL-8B-Instruct~\citep{bai2025qwen3vl}; the captioning model observes sampled rendered frames only and does not receive simulator metadata. Replays with rendering failures, broken synchronization, severe camera clipping, unstable simulation, or divergence not attributable to the intended physical edit are discarded.

\section{Dataset Utility for Gravity-Conditioned Generation}
\label{sec:utility-gen}

We evaluate PhysEditWorld as supervision for gravity-conditioned generation. Our goal is not to rank generative models comprehensively, but to test whether fine-tuning on matched gravity rollouts improves a model's response to explicit gravity conditions. This distinction matters because visually plausible generation can still fail physically: the camera may remain nearly static, falling may be delayed, or high-gravity rollouts may not accelerate more than low-gravity ones. PhysEditWorld makes these failures measurable by holding the scene, action sequence, camera policy, and initial state fixed while varying gravity.

\subsection{Experimental Setup}
\label{sec:setup}

\textbf{Data splits.} We construct training and evaluation splits at the \emph{replay-group} level rather than the clip level, so that no scene appears in both splits. The training set contains 1{,}530 (scene, action, gravity) tuples drawn from 9 scenes; the held-out set contains 170 tuples from the remaining 3 scenes. Within each replay group, all gravity variants are kept together to preserve the matched-comparison structure. We sample at most one clip per (scene, action, gravity) triplet to prevent near-duplicate frames from dominating the loss. Gravity multipliers are drawn from $\{0.05, 0.1, 0.5, 1.0, 2.0, 5.0, 20.0\}\times g_\oplus$.

\textbf{Baselines.}
We evaluate two settings.
For gravity-conditioned video generation, we compare zero-shot against
\textbf{Wan2.2-TI2V-5B}~\citep{wan2025wan} SFT version on our data.
% \textbf{Kling 3.0}~\citep{kuaishou2026kling3}, and
% \textbf{Seedance 2.0}~\citep{chen2026seedance2};
Kling 3.0 and Seedance 2.0 appear only in qualitative comparisons as they do not release training code,
while Wan2.2-TI2V-5B is additionally fine-tuned via LoRA~\cite{hu2022lora} on PhysEditWorld.
For action-conditioned first-person world modeling, we do quality study on 
\textbf{LingBot-World}~\citep{lingbotworld2026} fine-tuned on our data
and compare against the frozen \textbf{Matrix-Game 3.0}~\citep{matrixgame2026} and \textbf{LingBot-World}~\citep{lingbotworld2026}.

\textbf{Training protocol.} We apply LoRA with rank 128 on all attention projection matrices of both backbones, trained for 5 epochs with AdamW (learning rate $1\!\times\!10^{-4}$, batch size 8) on 8$\times$H100 GPUs. Inputs are 5-second clips at 30 FPS, $1280\!\times\!720$, with the gravity multiplier injected as a text token in the conditioning prompt (e.g., \texttt{"gravity: 0.25g"}). All other hyperparameters follow each backbone's released SFT recipe.

\textbf{Evaluation metric.} Since generated videos do not expose simulator states or ground-truth trajectories, we recover a per-frame camera trajectory with VGGT~\cite{wang2025vggt} and use its vertical-axis component as a one-dimensional fall-progress signal $q(t)$. We restrict evaluation to the pre-landing segment, detected via a simple progress threshold on smoothed $q(t)$, and discard clips with insufficient fall progress or unreliable landing detection. Because VGGT trajectories carry scale ambiguity, all speeds and accelerations are reported in normalized VGGT units, and only relative comparisons within a matched replay group are meaningful. Implementation details (smoothing window, threshold values, rejection rates) are provided.

For each reliable clip, we compute fall-axis speed $v(t)$ as the central-difference derivative of the smoothed $q(t)$ and fit a linear model:
\begin{equation}
    v(t) = at + b .
    \label{eq:linear-fit}
\end{equation}
The fitted slope $a$ serves as a normalized acceleration proxy, and $R^2$ measures how well the speed curve follows a linear acceleration pattern. To test whether motion ordering follows the requested gravity ordering, we compute pairwise gravity-acceleration alignment within each matched replay group:
\begin{equation}
    \mathrm{Align} = \frac{1}{|\mathcal{P}|} \sum_{(i,j)\in\mathcal{P}} \!\left[(g_i-g_j)(a_i-a_j)>0\right],
    \label{eq:alignment}
\end{equation}
where $\mathcal{P}=\{(i,j)\mid i<j,\ i,j\in\text{group}\}$ is the set of unordered pairs within a group, $g_i$ is the requested gravity, and $a_i$ is the fitted acceleration proxy. A perfectly gravity-faithful model achieves $\mathrm{Align}=1.0$; chance is $0.5$. 

% \subsection{Gravity-Conditioned Video Generation}
% \label{sec:wan-results}

% Table~\ref{tab:vggt-gravity-case} reports a representative matched case for Wan2.2-TI2V-5B before and after PhysEditWorld supervised fine-tuning. The zero-shot model is insensitive to the requested gravity: acceleration proxies are near-zero across all three settings, and the alignment of $33.3\%$ confirms that gravity ordering is not preserved. After PhysEditWorld SFT, the acceleration proxy increases monotonically with requested gravity, alignment reaches $100\%$, and mean $R^2$ rises from $0.066$ to $0.570$, indicating that the model now responds to gravity as a controllable variable.

\begin{figure}[t]
  \centering
  \includegraphics[width=0.95\linewidth]{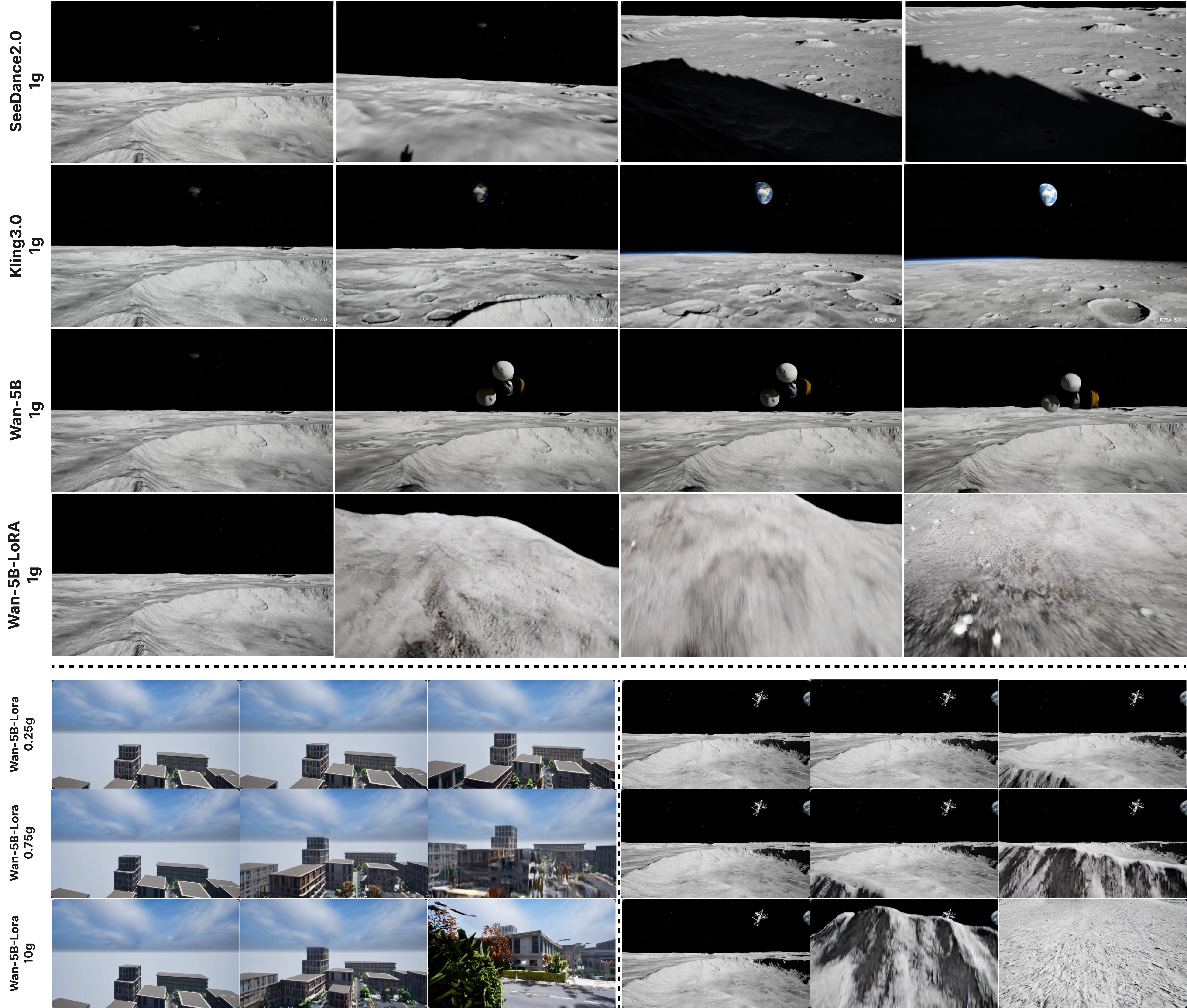}
  \caption{Qualitative results under a free-fall prompt.
    \emph{Top:} SeeDance2.0, Kling3.0, Wan2.2-TI2V-5B, and +PhysEditWorld LoRA at $1g$;
    baselines remain near-static while the LoRA model generates strong self-motion with motion blur.
    \emph{Bottom:} +PhysEditWorld LoRA at $0.25\!\times$/$0.75\!\times$/$10\!\times$ gravity;
    descent speed scales monotonically, confirming continuously controllable gravity.}
  \label{fig:qualitative-generation}
\end{figure}

% \begin{figure}
%     \centering
%     \includegraphics[width=1\linewidth]{figures/MoreQualitive.pdf}
%     \caption{Enter Caption}
%     \label{fig:placeholder}
% \end{figure}

\subsection{Gravity-Conditioned Video Generation / WorldModel}
\label{sec:wan-results}

Table~\ref{tab:vggt-gravity-case} reports a representative matched case for Wan2.2-TI2V-5B before and after PhysEditWorld supervised fine-tuning. The zero-shot model is insensitive to the requested gravity: acceleration proxies are near-zero across all three settings, and the alignment of $33.3\%$ confirms that gravity ordering is not preserved. After PhysEditWorld SFT, the acceleration proxy increases monotonically with requested gravity, alignment reaches $100\%$, and mean $R^2$ rises from $0.066$ to $0.570$, indicating that the model now responds to gravity as a controllable variable. Figure~\ref{fig:qualitative-generation} top shows a qualitative example under a $1g$ prompt: the zero-shot model produces minimal camera motion while the scene remains nearly static, whereas the LoRA-tuned model generates strong forward self-motion toward the lunar surface with visible motion blur.

\begin{table}[h]
  \centering
  \caption{VGGT-based gravity response analysis on a representative held-out matched replay group with three gravity settings ($0.25\!\times$, $0.75\!\times$, $10.0\!\times$). Accel. proxy denotes the fitted slope $a$ in $v(t)=at+b$ on the pre-landing segment, in normalized VGGT units; only \emph{within-row, within-group} comparisons are meaningful. Mean $R^2$ averages the linear-fit quality across the three settings.}
  \label{tab:vggt-gravity-case}
  \resizebox{\linewidth}{!}{%
  \begin{tabular}{lccccc}
    \toprule
    Model & Accel. @ $0.25\!\times$ & Accel. @ $0.75\!\times$ & Accel. @ $10.0\!\times$ & Mean $R^2$ $\uparrow$ & Gravity Align. $\uparrow$ \\
    \midrule
    Wan2.2-TI2V-5B (zero-shot) & 0.0002 & 0.0046 & 0.0009 & 0.066 & 33.3\% \\
    Wan2.2-TI2V-5B + PhysEditWorld SFT & 0.1766 & 0.3983 & 0.6214 & 0.570 & \textbf{100.0\%} \\
    \bottomrule
  \end{tabular}}
\end{table}

\label{sec:lingbot-results}
\begin{figure}[h]
  \centering
  \includegraphics[width=0.95\linewidth]{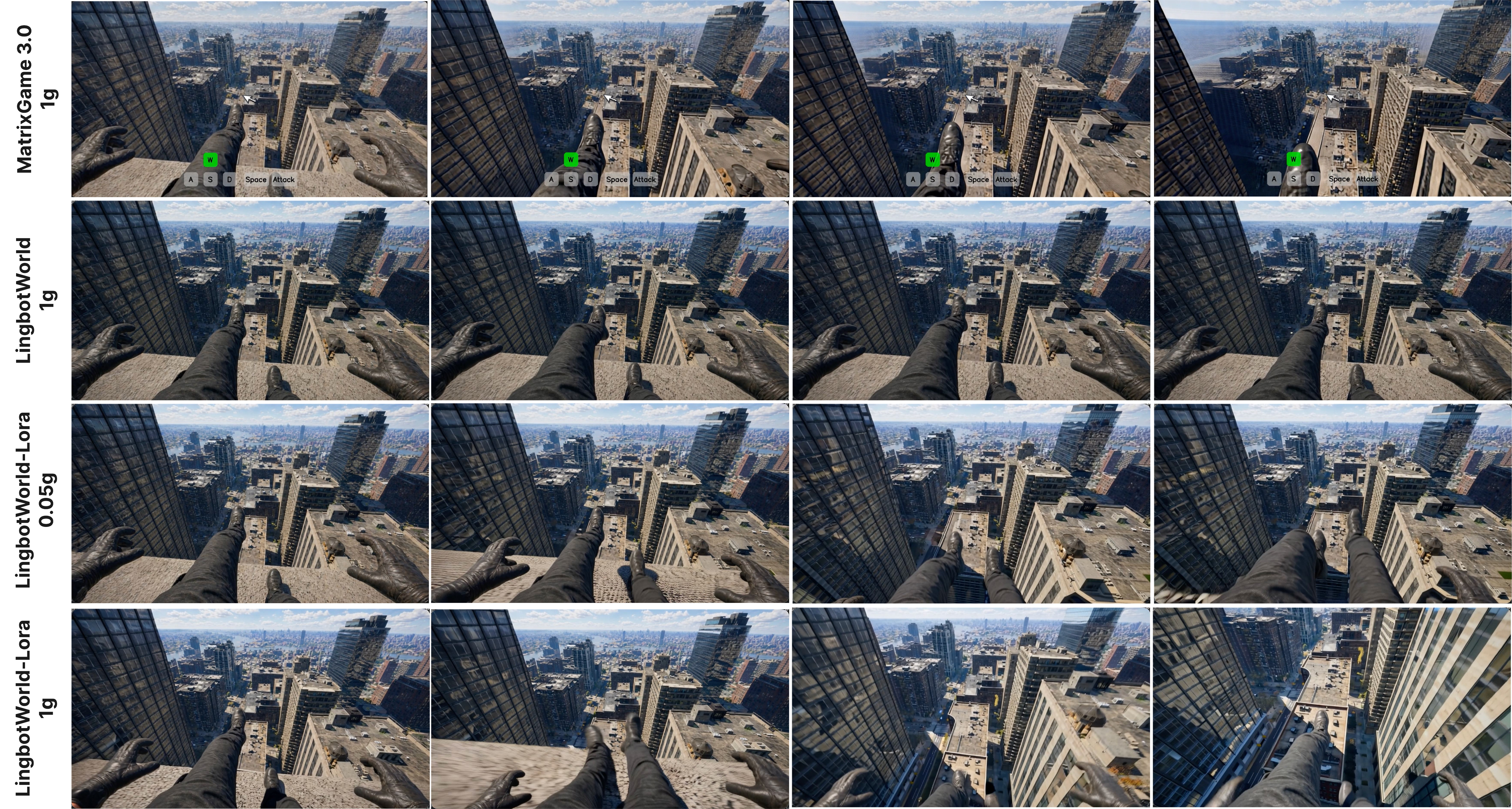}
    \caption{First-person world-model case study. First frame is generated by GPT-Image2. Baseline models (Matrix-Game 3.0, zero-shot LingBotWorld) remain near the platform edge and never enter free-fall under a $1g$ prompt. After PhysEditWorld LoRA-128 tuning, LingBotWorld generates platform departure and gravity-dependent downward self-motion.}
  \label{fig:wm-qualitative}
\end{figure}

\subsection{Action-Conditioned First-Person World Modeling}

We evaluate whether PhysEditWorld supervision transfers to action-conditioned first-person world models. We use a simple stress test: the \texttt{W} key is held continuously from a rooftop edge, where a gravity-faithful model should produce platform departure followed by gravity-dependent downward motion. As shown in Figure~\ref{fig:wm-qualitative}, both Matrix-Game 3.0 and zero-shot LingBotWorld remain near the platform edge and never enter free-fall. After PhysEditWorld LoRA tuning, LingBotWorld generates platform departure followed by downward self-motion that scales with the requested gravity. This demonstrates that the gravity-response failure is not limited to text-conditioned generation, and that PhysEditWorld provides effective supervision across both generation paradigms.

\section{Dataset Utility for Gravity-Aware VLMs}
\label{sec:utility-vlm}

\textbf{Experimental Setup.}
We evaluate Qwen3-VL-8B-Instruct~\cite{bai2025qwen3vl} on a held-out set of 170 rollouts drawn from the same stratified split described in \S\ref{sec:setup}.
The task requires the model to predict both the gravity class (low / normal / high) and the continuous gravity multiplier from a short gameplay video clip.

\textbf{Baseline.}
We first assess the zero-shot capability of Qwen3-VL-8B-Instruct \cite{bai2025qwen3vl} without any domain-specific fine-tuning, serving as a strong off-the-shelf VLM baseline.

\textbf{Training Protocol.}
We apply LoRA SFT on the remaining 1{,}530 training rollouts.
To discourage label memorization, gravity targets are jittered by $\pm10\%$ during training; evaluation is conducted against the original held-out labels.

\textbf{Metrics.}
We report class accuracy for the three-way gravity classification, and mean absolute error (MAE), median absolute percentage error (Median APE), and within-10\% rate for the continuous gravity multiplier prediction.

\begin{table}[H]
  \centering
  \caption{Gravity-aware prediction on held-out rollouts. +SFT denotes PhysEditWorld LoRA tuning.}
  \label{tab:vlm-utility}
  \resizebox{\linewidth}{!}{%
  \begin{tabular}{lcccc}
    \toprule
    Model & Class Acc. (\%) $\uparrow$ & MAE $\downarrow$ & Median APE (\%) $\downarrow$ & Within 10\% $\uparrow$ \\
    \midrule
    Qwen3-VL-8B\cite{bai2025qwen3vl}            & 24.71 & 5.4573 & 95.00 &  9.48 \\
    Qwen3-VL-8B + SFT      & 95.29 & 0.8305 &  6.22 & 90.59 \\
    \bottomrule
  \end{tabular}}
\end{table}

As shown in Table~\ref{tab:vlm-utility}, the zero-shot model achieves a class accuracy of 24.71\%---below the 33.3\% random baseline for a three-way classification task---with a median APE of 95.00\% and a within-10\% rate of only 9.48\%, confirming that gravity magnitude is not reliably recoverable from video without targeted supervision.
After PhysEditWorld LoRA SFT, class accuracy improves to 95.29\% and median APE drops to 6.22\%, with 90.59\% of predictions falling within 10\% of the true gravity multiplier.
These results demonstrate that the physics-varied rollouts in PhysEditWorld provide effective supervision signal for fine-grained, gravity-aware video-language understanding.

\section{Conclusion}

PhysEditWorld introduces editable game-world modeling as a controlled physical-intervention problem. Instead of asking only whether a generated rollout looks plausible, it asks whether the same authored scenario evolves consistently when a physical rule is changed. By replaying matched interactions under explicit gravity configurations, PhysEditWorld separates visual realism from physical controllability and reveals failure modes that standard video-quality metrics can overlook. The current release focuses on gravity as a measurable and widely supported first step, covering effects such as airtime, fall speed, jump arcs, and landing dynamics. In future releases, we plan to extend the dataset and pipeline to additional editable physical attributes, such as friction, drag, restitution, wind, and object-level physical parameters, further supporting controllable neural game engines that can be edited as authored simulations rather than only imitated as videos.

\bibliographystyle{unsrtnat}
\bibliography{references}

\begin{thebibliography}{10}
\providecommand{\url}[1]{\texttt{#1}}
\providecommand{\urlprefix}{URL }
\providecommand{\doi}[1]{https://doi.org/#1}

\bibitem{alonso2024diamond}
Alonso, E., Jelley, A., Micheli, V., Kanervisto, A., Storkey, A., Pearce, T., Fleuret, F.: Diffusion for world modeling: Visual details matter in {Atari}. In: Advances in Neural Information Processing Systems (2024)

\bibitem{Qwen3-VL}
Bai, S., Cai, Y., Chen, R., Chen, K., Chen, X., Cheng, Z., Deng, L., Ding, W., Gao, C., Ge, C., Ge, W., Guo, Z., Huang, Q., Huang, J., Huang, F., Hui, B., Jiang, S., Li, Z., Li, M., Li, M., Li, K., Lin, Z., Lin, J., Liu, X., Liu, J., Liu, C., Liu, Y., Liu, D., Liu, S., Lu, D., Luo, R., Lv, C., Men, R., Meng, L., Ren, X., Ren, X., Song, S., Sun, Y., Tang, J., Tu, J., Wan, J., Wang, P., Wang, P., Wang, Q., Wang, Y., Xie, T., Xu, Y., Xu, H., Xu, J., Yang, Z., Yang, M., Yang, J., Yang, A., Yu, B., Zhang, F., Zhang, H., Zhang, X., Zheng, B., Zhong, H., Zhou, J., Zhou, F., Zhou, J., Zhu, Y., Zhu, K.: Qwen3-vl technical report. arXiv preprint arXiv:2511.21631  (2025)

\bibitem{bai2025qwen3vl}
Bai, S., Cai, Y., Chen, R., Chen, K., Chen, X., Cheng, Z., Deng, L., Ding, W., Gao, C., Ge, C., Ge, W., Guo, Z., Huang, Q., Huang, J., Huang, F., Hui, B., Jiang, S., Li, Z., Li, M., Li, M., Li, K., Lin, Z., Lin, J., Liu, X., Liu, J., Liu, C., Liu, Y., Liu, D., Liu, S., Lu, D., Luo, R., Lv, C., Men, R., Meng, L., Ren, X., Ren, X., Song, S., Sun, Y., Tang, J., Tu, J., Wan, J., Wang, P., Wang, P., Wang, Q., Wang, Y., Xie, T., Xu, Y., Xu, H., Xu, J., Yang, Z., Yang, M., Yang, J., Yang, A., Yu, B., Zhang, F., Zhang, H., Zhang, X., Zheng, B., Zhong, H., Zhou, J., Zhou, F., Zhou, J., Zhu, Y., Zhu, K.: {Qwen3-VL} technical report (2025)

\bibitem{baker2022vpt}
Baker, B., Akkaya, I., Zhokhov, P., Huizinga, J., Tang, J., Ecoffet, A., Houghton, B., Sampedro, R., Clune, J.: Video pretraining ({VPT}): Learning to act by watching unlabeled online videos. In: Advances in Neural Information Processing Systems (2022), \url{https://arxiv.org/abs/2206.11795}

\bibitem{bakhtin2019phyre}
Bakhtin, A., van~der Maaten, L., Johnson, J., Gustafson, L., Girshick, R.: {PHYRE}: A new benchmark for physical reasoning. In: Advances in Neural Information Processing Systems (2019)

\bibitem{bansal2025videophy}
Bansal, H., Lin, Z., Xie, T., Zong, Z., Yarom, M., Bitton, Y., Jiang, C., Sun, Y., Chang, K.W., Grover, A.: {VideoPhy}: Evaluating physical commonsense for video generation. In: International Conference on Learning Representations (2025)

\bibitem{baradel2020cophy}
Baradel, F., Neverova, N., Mille, J., Mori, G., Wolf, C.: {CoPhy}: Counterfactual learning of physical dynamics. In: International Conference on Learning Representations (2020), \url{https://openreview.net/forum?id=SkeyppEFvS}

\bibitem{bear2021physion}
Bear, D.M., Wang, E., Mrowca, D., Binder, F.J., et~al.: Physion: Evaluating physical prediction from vision in humans and machines. In: Advances in Neural Information Processing Systems Datasets and Benchmarks Track (2021)

\bibitem{bellemare2013ale}
Bellemare, M.G., Naddaf, Y., Veness, J., Bowling, M.: The arcade learning environment: An evaluation platform for general agents. Journal of Artificial Intelligence Research  \textbf{47},  253--279 (2013)

\bibitem{bordes2025intphys2}
Bordes, F., Garrido, Q., Kao, J.T., Williams, A., Rabbat, M., Dupoux, E.: {IntPhys} 2: Benchmarking intuitive physics understanding in complex synthetic environments. arXiv preprint arXiv:2506.09849  (2025), \url{https://arxiv.org/abs/2506.09849}

\bibitem{bruce2024genie}
Bruce, J., Dennis, M., Edwards, A., Parker-Holder, J., Shi, Y., Hughes, E., Lai, M., Mavalankar, A., Steigerwald, R., Apps, C., Aytar, Y., Bechtle, S., Behbahani, F., Chan, S.C.Y., Heess, N., Gonzalez, L., Osindero, S., Ozair, S., Reed, S., Zhang, J., Zolna, K., Clune, J., de~Freitas, N., Singh, S., Rockt{\"a}schel, T.: Genie: Generative interactive environments. In: Proceedings of the 41st International Conference on Machine Learning. Proceedings of Machine Learning Research, vol.~235, pp. 4603--4623. PMLR (2024)

\bibitem{che2024gamegenx}
Che, H., He, X., Liu, Q., Jin, C., Chen, H.: Gamegen-x: Interactive open-world game video generation. In: International Conference on Learning Representations (2025)

\bibitem{chen2025interactiveworldmodels}
Chen, T., Hu, X., Ding, Z., Jin, C.: Learning world models for interactive video generation. arXiv preprint arXiv:2505.21996  (2025)

\bibitem{chen2022comphy}
Chen, Z., Yi, K., Li, Y., Ding, M., Torralba, A., Tenenbaum, J.B., Gan, C.: {ComPhy}: Compositional physical reasoning of objects and events from videos. In: International Conference on Learning Representations (2022), \url{https://openreview.net/forum?id=PgNEYaIc81Q}

\bibitem{cobbe2020procgen}
Cobbe, K., Hesse, C., Hilton, J., Schulman, J.: Leveraging procedural generation to benchmark reinforcement learning. In: International Conference on Machine Learning (2020)

\bibitem{quevedo2024oasis}
{Decart}, Quevedo, J., McIntyre, Q., Campbell, S., Chen, X., Wachen, R.: Oasis: A universe in a transformer. Project page (2024), \url{https://oasis-model.github.io/}

\bibitem{deitke2020robothor}
Deitke, M., Han, W., Herrasti, A., Kembhavi, A., Kolve, E., Mottaghi, R., Salvador, J., Schwenk, D., VanderBilt, E., Wallingford, M., Weihs, L., Yatskar, M., Farhadi, A.: {RoboTHOR}: An open simulation-to-real embodied {AI} platform. In: Proceedings of the IEEE/CVF Conference on Computer Vision and Pattern Recognition (2020), \url{https://arxiv.org/abs/2004.06799}

\bibitem{deitke2022procthor}
Deitke, M., VanderBilt, E., Herrasti, A., Weihs, L., Salvador, J., Ehsani, K., Han, W., Kolve, E., Farhadi, A., Kembhavi, A., Mottaghi, R.: {ProcTHOR}: Large-scale embodied {AI} using procedural generation. In: Advances in Neural Information Processing Systems (2022), \url{https://proceedings.neurips.cc/paper_files/paper/2022/hash/27c546ab1e4f1d7d638e6a8dfbad9a07-Abstract-Conference.html}, outstanding Paper Award

\bibitem{dosovitskiy2017carla}
Dosovitskiy, A., Ros, G., Codevilla, F., Lopez, A., Koltun, V.: {CARLA}: An open urban driving simulator. In: Conference on Robot Learning (2017)

\bibitem{duan2025worldscore}
Duan, H., Yu, H.X., Chen, S., Fei-Fei, L., Wu, J.: Worldscore: A unified evaluation benchmark for world generation. In: Proceedings of the IEEE/CVF International Conference on Computer Vision (2025)

\bibitem{fan2022minedojo}
Fan, L., Wang, G., Jiang, Y., Mandlekar, A., Yang, Y., Zhu, H., Tang, A., Huang, D.A., Zhu, Y., Anandkumar, A.: {MineDojo}: Building open-ended embodied agents with internet-scale knowledge. In: Advances in Neural Information Processing Systems Datasets and Benchmarks Track (2022), \url{https://openreview.net/forum?id=rc8o_j8I8PX}

\bibitem{foss2025causalvqa}
Foss, A., Evans, C., Mitts, S., Sinha, K., Rizvi, A., Kao, J.T.: {CausalVQA}: A physically grounded causal reasoning benchmark for video models. arXiv preprint arXiv:2506.09943  (2025)

\bibitem{gan2021tdw}
Gan, C., Schwartz, J., Alter, S., Mrowca, D., Schrimpf, M., Traer, J., De~Freitas, J., Kubilius, J., Bhandwaldar, A., Haber, N., et~al.: {ThreeDWorld}: A platform for interactive multi-modal physical simulation. In: Advances in Neural Information Processing Systems Datasets and Benchmarks Track (2021)

\bibitem{gillman2025forceprompting}
Gillman, N., Herrmann, C., Freeman, M., Aggarwal, D., Luo, E., Sun, D., Sun, C.: Force prompting: Video generation models can learn and generalize physics-based control signals. In: Advances in Neural Information Processing Systems (2025), \url{https://arxiv.org/abs/2505.19386}

\bibitem{girdhar2020cater}
Girdhar, R., Ramanan, D.: {CATER}: A diagnostic dataset for compositional actions and {TE}mporal reasoning. In: International Conference on Learning Representations (2020), \url{https://arxiv.org/abs/1910.04744}

\bibitem{greff2022kubric}
Greff, K., Belletti, F., Beyer, L., Doersch, C., Du, Y., Duckworth, D., Fleet, D.J., Gnanapragasam, D., Golemo, F., Herrmann, C., et~al.: Kubric: A scalable dataset generator. In: Proceedings of the IEEE/CVF Conference on Computer Vision and Pattern Recognition (2022)

\bibitem{groth2018shapestacks}
Groth, O., Fuchs, F.B., Posner, I., Vedaldi, A.: {ShapeStacks}: Learning vision-based physical intuition for generalised object stacking. In: European Conference on Computer Vision (2018), \url{https://arxiv.org/abs/1804.08018}

\bibitem{gu2023maniskill2}
Gu, J., Xiang, F., Li, X., Ling, Z., Liu, X., Mu, T., Tang, Y., Tao, S., Wei, X., Yao, Y., Yuan, X., Xie, P., Huang, Z., Chen, R., Su, H.: {ManiSkill2}: A unified benchmark for generalizable manipulation skills. In: International Conference on Learning Representations (2023), \url{https://arxiv.org/abs/2302.04659}

\bibitem{guo2025mineworld}
Guo, J., Ye, Y., He, T., Wu, H., Jiang, Y., Pearce, T., Bian, J.: Mineworld: A real-time and open-source interactive world model on {Minecraft}. arXiv preprint arXiv:2504.08388  (2025)

\bibitem{guss2019minerl}
Guss, W.H., Houghton, B., Topin, N., Wang, P., Codel, C., Veloso, M., Salakhutdinov, R.: {MineRL}: A large-scale dataset of {Minecraft} demonstrations. In: International Joint Conference on Artificial Intelligence (2019)

\bibitem{hu2026identityconsistentvideogenerationlarge}
Hu, B., Qi, Z., Huang, G., Xu, Z., Zhang, R., Ye, C., Zhou, J., Li, X., Wang, J.: Identity-consistent video generation under large facial-angle variations (2026), \url{https://arxiv.org/abs/2603.21299}

\bibitem{hu2022lora}
Hu, E.J., Shen, Y., Wallis, P., Allen-Zhu, Z., Li, Y., Wang, S., Chen, W.: {LoRA}: Low-rank adaptation of large language models. In: International Conference on Learning Representations (2022)

\bibitem{huang2025dc}
Huang, T., Zhang, Z., Zhang, R., Zhao, Y.: Dc-scene: Data-centric learning for 3d scene understanding. arXiv preprint arXiv:2505.15232  (2025)

\bibitem{huang2024vbench}
Huang, Z., He, Y., Yu, J., et~al.: {VBench}: Comprehensive benchmark suite for video generative models. In: Proceedings of the IEEE/CVF Conference on Computer Vision and Pattern Recognition (2024)

\bibitem{james2020rlbench}
James, S., Ma, Z., Arrojo, D.R., Davison, A.J.: {RLBench}: The robot learning benchmark and learning environment. IEEE Robotics and Automation Letters  \textbf{5}(2),  3019--3026 (2020), \url{https://arxiv.org/abs/1909.12271}

\bibitem{kim2020gamegan}
Kim, S.W., Zhou, Y., Philion, J., Torralba, A., Fidler, S.: Learning to simulate dynamic environments with {GameGAN}. In: Proceedings of the IEEE/CVF Conference on Computer Vision and Pattern Recognition (2020)

\bibitem{kolve2017ai2thor}
Kolve, E., Mottaghi, R., Han, W., VanderBilt, E., Weihs, L., Herrasti, A., Deitke, M., Ehsani, K., Gordon, D., Zhu, Y., et~al.: {AI2-THOR}: An interactive 3d environment for visual {AI}. arXiv preprint arXiv:1712.05474  (2017)

\bibitem{kuaishou2026kling3}
{Kuaishou Technology}: {Kling AI} launches 3.0 model, ushering in an era where everyone can be a director. Company news release (2026), published February 5, 2026

\bibitem{le2025newtonrewards}
Le, M.Q., Zhu, Y., Kalogeiton, V., Samaras, D.: What about gravity in video generation? post-training newton's laws with verifiable rewards. arXiv preprint arXiv:2512.00425  (2025)

\bibitem{li2022igibson2}
Li, C., Xia, F., Mart{\'i}n-Mart{\'i}n, R., Lingelbach, M., Srivastava, S., Shen, B., Vainio, K.E., Gokmen, C., Dharan, G., Jain, T., Kurenkov, A., Liu, K., Gweon, H., Wu, J., Fei-Fei, L., Savarese, S.: {iGibson} 2.0: Object-centric simulation for robot learning of everyday household tasks. In: Proceedings of the 5th Conference on Robot Learning. Proceedings of Machine Learning Research, vol.~164, pp. 455--465. PMLR (2022), \url{https://proceedings.mlr.press/v164/li22b.html}

\bibitem{li2024behavior1k}
Li, C., Zhang, R., Wong, J., Gokmen, C., Srivastava, S., Mart{\'i}n-Mart{\'i}n, R., Wang, C., Levine, G., Ai, W., Martinez, B., et~al.: {BEHAVIOR}-1k: A human-centered, embodied {AI} benchmark with 1,000 everyday activities and realistic simulation. arXiv preprint arXiv:2403.09227  (2024)

\bibitem{li2025pisa}
Li, C., Michel, O., Pan, X., Liu, S., Roberts, M., Xie, S.: {PISA} experiments: Exploring physics post-training for video diffusion models by watching stuff drop. In: International Conference on Machine Learning (2025), \url{https://arxiv.org/abs/2503.09595}

\bibitem{li2025quantiphy}
Li, P., Xiang, T., Mao, E., Wei, S., Chen, X., Masood, A., Fei-Fei, L., Adeli, E.: {QuantiPhy}: A quantitative benchmark evaluating physical reasoning abilities of vision-language models. arXiv preprint arXiv:2512.19526  (2025)

\bibitem{li2024iphyre}
Li, S., Wu, K., Zhang, C., Zhu, Y.: {I-PHYRE}: Interactive physical reasoning. In: International Conference on Learning Representations (2024), \url{https://openreview.net/forum?id=1bbPQShCT2}

\bibitem{li2025sekai}
Li, Z., Li, C., Mao, X., Lin, S., Li, M., Zhao, S., Xu, Z., Li, X., Feng, Y., Sun, J., et~al.: Sekai: A video dataset towards world exploration. In: Advances in Neural Information Processing Systems Datasets and Benchmarks Track (2025)

\bibitem{li2026wildworld}
Li, Z., Meng, Z., Shi, S., Peng, W., Wu, Y., Zheng, B., Li, C., Zhang, K.: Wildworld: A large-scale dataset for dynamic world modeling with actions and explicit state toward generative {ARPG}. arXiv preprint arXiv:2603.23497  (2026), \url{https://arxiv.org/abs/2603.23497}

\bibitem{liu2024physgen}
Liu, S., Ren, Z., Gupta, S., Wang, S.: {PhysGen}: Rigid-body physics-grounded image-to-video generation. In: European Conference on Computer Vision (2024). \doi{10.1007/978-3-031-73007-8_21}, \url{https://arxiv.org/abs/2409.18964}

\bibitem{makoviychuk2021isaacgym}
Makoviychuk, V., Wawrzyniak, L., Guo, Y., Lu, M., Storey, K., Macklin, M., Hoeller, D., Rudin, N., Allshire, A., Handa, A., State, G.: {Isaac Gym}: High performance {GPU}-based physics simulation for robot learning. In: Advances in Neural Information Processing Systems Datasets and Benchmarks Track (2021), \url{https://arxiv.org/abs/2108.10470}

\bibitem{mao2025yume}
Mao, X., Lin, S., Li, Z., Li, C., Peng, W., He, T., Pang, J., Chi, M., Qiao, Y., Zhang, K.: Yume: An interactive world generation model. arXiv preprint arXiv:2507.17744  (2025)

\bibitem{menapace2022playableenv}
Menapace, W., Lathuili{\`e}re, S., Siarohin, A., Theobalt, C., Tulyakov, S., Golyanik, V., Ricci, E.: Playable environments: Video manipulation in space and time. In: Proceedings of the IEEE/CVF Conference on Computer Vision and Pattern Recognition. pp. 3584--3593 (2022), \url{https://arxiv.org/abs/2203.01914}

\bibitem{menapace2021pvg}
Menapace, W., Lathuiliere, S., Tulyakov, S., Siarohin, A., Ricci, E.: Playable video generation. In: Proceedings of the IEEE/CVF Conference on Computer Vision and Pattern Recognition (2021)

\bibitem{menapace2024pgm}
Menapace, W., Siarohin, A., Lathuiliere, S., Achlioptas, P., Golyanik, V., Tulyakov, S., Ricci, E.: Promptable game models: Text-guided game simulation via masked diffusion models. ACM Transactions on Graphics  (2024)

\bibitem{meng2025phygenbench}
Meng, F., Liao, J., Tan, X., Shao, W., Lu, Q., Zhang, K., Cheng, Y., Li, D., Qiao, Y., Luo, P.: Towards world simulator: Crafting physical commonsense-based benchmark for video generation. In: International Conference on Machine Learning (2025)

\bibitem{motamed2025physicsiq}
Motamed, S., Culp, L., Swersky, K., Jaini, P., Geirhos, R.: Do generative video models understand physical principles? In: Proceedings of the IEEE/CVF Winter Conference on Applications of Computer Vision. pp. 948--958 (March 2026), \url{https://arxiv.org/abs/2501.09038}

\bibitem{narayanan2026phyco}
Narayanan, S., Jiang, Z., Narasimhan, S., Chandraker, M.: Phyco: Learning controllable physical priors for generative motion (2026), \url{https://arxiv.org/abs/2604.28169}

\bibitem{patel2022crippvqa}
Patel, M., Gokhale, T., Baral, C., Yang, Y.: {CRIPP}-{VQA}: Counterfactual reasoning about implicit physical properties via video question answering. In: Proceedings of the 2022 Conference on Empirical Methods in Natural Language Processing. pp. 9856--9870. Association for Computational Linguistics, Abu Dhabi, United Arab Emirates (2022). \doi{10.18653/v1/2022.emnlp-main.670}, \url{https://aclanthology.org/2022.emnlp-main.670/}

\bibitem{puig2018virtualhome}
Puig, X., Ra, K., Boben, M., Li, J., Wang, T., Fidler, S., Torralba, A.: {VirtualHome}: Simulating household activities via programs. In: Proceedings of the IEEE/CVF Conference on Computer Vision and Pattern Recognition. pp. 8494--8502 (2018), \url{https://arxiv.org/abs/1806.07011}

\bibitem{qiu2016unrealcv}
Qiu, W., Yuille, A.: {UnrealCV}: Connecting computer vision to unreal engine. arXiv preprint arXiv:1609.01326  (2016), \url{https://arxiv.org/abs/1609.01326}

\bibitem{riochet2022intphys}
Riochet, R., Castro, M.Y., Bernard, M., Lerer, A., Fergus, R., Izard, V., Dupoux, E.: {IntPhys} 2019: A benchmark for visual intuitive physics understanding. IEEE Transactions on Pattern Analysis and Machine Intelligence  \textbf{44}(9),  5016--5025 (2022)

\bibitem{lingbotworld2026}
{Robbyant Team}, Gao, Z., Wang, Q., Zeng, Y., Zhu, J., Cheng, K.L., Li, Y., Wang, H., Xu, Y., Ma, S., Chen, Y., Liu, J., Cheng, Y., Yao, Y., Zhu, J., Meng, Y., Zheng, K., Bai, Q., Chen, J., Shen, Z., Yu, Y., Zhu, X., Shen, Y., Ouyang, H.: Advancing open-source world models. arXiv preprint arXiv:2601.20540  (2026)

\bibitem{savva2019habitat}
Savva, M., Kadian, A., Maksymets, O., Zhao, Y., Wijmans, E., Jain, B., Straub, J., Liu, J., Koltun, V., Malik, J., et~al.: Habitat: A platform for embodied {AI} research. In: Proceedings of the IEEE/CVF International Conference on Computer Vision (2019)

\bibitem{smith2019adept}
Smith, K.A., Mei, L., Yao, S., Wu, J., Spelke, E.S., Tenenbaum, J.B., Ullman, T.D.: Modeling expectation violation in intuitive physics with coarse probabilistic object representations. In: Advances in Neural Information Processing Systems (2019), \url{https://proceedings.neurips.cc/paper/2019/hash/e88f243bf341ded9b4ced444795c3f17-Abstract.html}

\bibitem{taesiri2022clipgamephysics}
Taesiri, M.R., Macklon, F., Bezemer, C.P.: {CLIP} meets {GamePhysics}: Towards bug identification in gameplay videos using zero-shot transfer learning. In: 2022 IEEE/ACM 19th International Conference on Mining Software Repositories. pp. 270--281 (2022). \doi{10.1145/3524842.3528438}, \url{https://arxiv.org/abs/2203.11096}

\bibitem{chen2026seedance2}
{Team Seedance}, Chen, D., Chen, L., Chen, X., Chen, Y., Chen, Z., Chen, Z., Cheng, F., Cheng, T., Cheng, Y., et~al.: Seedance 2.0: Advancing video generation for world complexity. arXiv preprint arXiv:2604.14148  (2026)

\bibitem{valevski2024gamengen}
Valevski, D., Leviathan, Y., Arar, M., Fruchter, S.: Diffusion models are real-time game engines. In: International Conference on Learning Representations (2025), \url{https://arxiv.org/abs/2408.14837}

\bibitem{wan2025wanopenadvancedlargescale}
Wan, T., Wang, A., Ai, B., Wen, B., Mao, C., Xie, C.W., Chen, D., Yu, F., Zhao, H., Yang, J., Zeng, J., Wang, J., Zhang, J., Zhou, J., Wang, J., Chen, J., Zhu, K., Zhao, K., Yan, K., Huang, L., Feng, M., Zhang, N., Li, P., Wu, P., Chu, R., Feng, R., Zhang, S., Sun, S., Fang, T., Wang, T., Gui, T., Weng, T., Shen, T., Lin, W., Wang, W., Wang, W., Zhou, W., Wang, W., Shen, W., Yu, W., Shi, X., Huang, X., Xu, X., Kou, Y., Lv, Y., Li, Y., Liu, Y., Wang, Y., Zhang, Y., Huang, Y., Li, Y., Wu, Y., Liu, Y., Pan, Y., Zheng, Y., Hong, Y., Shi, Y., Feng, Y., Jiang, Z., Han, Z., Wu, Z.F., Liu, Z.: Wan: Open and advanced large-scale video generative models (2025), \url{https://arxiv.org/abs/2503.20314}

\bibitem{wan2025wan}
{Wan Team}, Wang, A., Ai, B., Wen, B., Mao, C., Xie, C.W., Chen, D., Yu, F., Zhao, H., Yang, J., Zeng, J., Wang, J., Zhang, J., Zhou, J., et~al.: Wan: Open and advanced large-scale video generative models. arXiv preprint arXiv:2503.20314  (2025)

\bibitem{wang2025physctrl}
Wang, C., Chen, C., Huang, Y., Dou, Z., Liu, Y., Gu, J., Liu, L.: {PhysCtrl}: Generative physics for controllable and physics-grounded video generation. arXiv preprint arXiv:2509.20358  (2025)

\bibitem{wang2025vggt}
Wang, J., Chen, M., Karaev, N., Vedaldi, A., Rupprecht, C., Novotny, D.: {VGGT}: Visual geometry grounded transformer. In: Proceedings of the IEEE/CVF Conference on Computer Vision and Pattern Recognition (2025)

\bibitem{wang2025visualactionprompts}
Wang, Y., Wen, C., Guo, H., Peng, S., Qin, M., Bao, H., Zhou, X., Hu, R.: Precise action-to-video generation through visual action prompts. In: Proceedings of the IEEE/CVF International Conference on Computer Vision (2025), \url{https://arxiv.org/abs/2508.13104}

\bibitem{matrixgame2026}
Wang, Z., Liu, Z., Li, J., Huang, K., Xu, B., Kang, F., An, M., Wang, P., Jiang, B., Wei, Y., Xietian, Y., Pei, J., Hu, L., Jiang, B., Xue, H., Wang, Z., Sun, H., Li, W., Ouyang, W., He, X., Liu, Y., Li, Y., Zhou, Y.: Matrix-game 3.0: Real-time and streaming interactive world model with long-horizon memory. arXiv preprint arXiv:2604.08995  (2026)

\bibitem{wu2026multiworld}
Wu, H., Yu, J., Zou, Y., Liu, X.: Multiworld: Scalable multi-agent multi-view video world models. arXiv preprint arXiv:2604.18564  (2026)

\bibitem{xia2018gibson}
Xia, F., Zamir, A.R., He, Z., Sax, A., Malik, J., Savarese, S.: Gibson env: Real-world perception for embodied agents. In: Proceedings of the IEEE Conference on Computer Vision and Pattern Recognition (2018), \url{https://openaccess.thecvf.com/content_cvpr_2018/html/Xia_Gibson_Env_Real-World_CVPR_2018_paper.html}

\bibitem{xiang2020sapien}
Xiang, F., Qin, Y., Mo, K., Xia, Y., Zhu, H., Liu, F., Liu, M., Jiang, H., Yuan, Y., Wang, H., Yi, L., Chang, A.X., Guibas, L.J., Su, H.: {SAPIEN}: A simulated part-based interactive environment. In: Proceedings of the IEEE/CVF Conference on Computer Vision and Pattern Recognition (2020), \url{https://arxiv.org/abs/2003.08515}

\bibitem{xu2026eponav2}
Xu, J., Zhong, Z., Shu, Z., Jia, M., Li, M., Bian, J.W., Zhang, Q., Zhang, K., Xie, J., Yang, J., et~al.: Eponav2: Driving world model with comprehensive future reasoning. arXiv preprint arXiv:2605.14696  (2026)

\bibitem{yi2020clevrer}
Yi, K., Gan, C., Li, Y., Kohli, P., Wu, J., Torralba, A., Tenenbaum, J.B.: {CLEVRER}: Collision events for video representation and reasoning. In: International Conference on Learning Representations (2020)

\bibitem{yu2025gamefactory}
Yu, J., Qin, Y., Wang, X., Wan, P., Zhang, D., Liu, X.: Gamefactory: Creating new games with generative interactive videos. In: Proceedings of the IEEE/CVF International Conference on Computer Vision (2025)

\bibitem{yuan2026newtongen}
Yuan, Y., Wang, X., Wickremasinghe, T., Nadir, Z., Ma, B., Chan, S.H.: {NewtonGen}: Physics-consistent and controllable text-to-video generation via neural newtonian dynamics. In: International Conference on Learning Representations (2026), \url{https://openreview.net/forum?id=rJ6N6sunaU}

\bibitem{zhang2025physchoreo}
Zhang, H., Huang, T., Wan, Z., Jin, X., Zhang, H., Li, H., Zuo, W.: {PhysChoreo}: Physics-controllable video generation with part-aware semantic grounding. arXiv preprint arXiv:2511.20562  (2025)

\bibitem{zhang2026robostereodualtower4dembodied}
Zhang, R., Chen, G., Xu, Z., Liu, Z., Zhong, Z., Zhang, M., Zhou, J., Li, X.: Robostereo: Dual-tower 4d embodied world models for unified policy optimization (2026), \url{https://arxiv.org/abs/2603.12639}

\bibitem{zhang2026mindvhierarchicalworldmodel}
Zhang, R., Zhang, M., Zhou, J., Liu, X., Xu, Z., Zhong, Z., Yan, P., Luo, H., Li, X.: Mind-v: Hierarchical world model for long-horizon robotic manipulation with rl-based physical alignment (2026), \url{https://arxiv.org/abs/2512.06628}

\bibitem{zheng2025vbench2}
Zheng, D., Huang, Z., Liu, H., Zou, K., He, Y., Zhang, F., Gu, L., Zhang, Y., He, J., Zheng, W.S., Qiao, Y., Liu, Z.: {VBench}-2.0: Advancing video generation benchmark suite for intrinsic faithfulness. arXiv preprint arXiv:2503.21755  (2025), \url{https://arxiv.org/abs/2503.21755}

\bibitem{zheng2024contphy}
Zheng, Z., Yan, X., Chen, Z., Wang, J., Lim, Q.Z.E., Tenenbaum, J.B., Gan, C.: {ContPhy}: Continuum physical concept learning and reasoning from videos. In: Proceedings of the 41st International Conference on Machine Learning. Proceedings of Machine Learning Research, vol.~235, pp. 61526--61558. PMLR (2024), \url{https://proceedings.mlr.press/v235/zheng24l.html}

\bibitem{zhong2025unrealzoo}
Zhong, F., Wu, K., Wang, C., Chen, H., Ci, H., Li, Z., Wang, Y.: {UnrealZoo}: Enriching photo-realistic virtual worlds for embodied {AI}. In: Proceedings of the IEEE/CVF International Conference on Computer Vision (2025), \url{https://arxiv.org/abs/2412.20977}, highlight

\bibitem{zhou2026physinone}
Zhou, S., Wang, H., Cheng, H., Li, J., Wang, D., Jiang, J., Jin, Y., Huang, J., Mao, S., Liu, S., et~al.: Physinone: Visual physics learning and reasoning in one suite. arXiv preprint arXiv:2604.09415  (2026)

\end{thebibliography}

% \bibliographystyle{plainnat}
% \bibliography{references}

\appendix

\clearpage

% \section{Additional Dataset Details}
% \label{app:dataset-details}
\section{Asset Library, Character Setup, and Capture Configuration}

\subsection{Asset Library}

PhysEditWorld is built from a manually curated UE5 asset library designed to cover diverse gravity-sensitive interaction scenarios. The current asset collection contains 12 representative scene environments, including polar research bases, indoor sports courts, lunar and Martian surfaces, cyberpunk city blocks, classrooms, laboratories, forest valleys, research camps, bowling alleys, theaters, and modern urban streets. These assets span indoor, outdoor, planetary, urban, sports, and object-interaction settings, allowing the dataset to expose different forms of gravity-dependent behavior such as free fall, jump arcs, landing timing, object displacement, and contact response. Artists manually inspect each scene for visual fidelity, collision quality, interaction suitability, and physical stability before it is included in the replay pipeline.

In addition to scene assets, PhysEditWorld includes three character assets with compatible animation and retargeting setups. The character system uses a UE5 animation stack based on motion matching, pose search, chooser-driven animation selection, blend stacks, orientation warping, and animation warping. To support low-gravity scenarios, we implement a low-gravity animation tuner that adjusts airborne-state detection and animation playback rate according to the gravity multiplier. This prevents low-gravity clips from using visually implausible Earth-gravity animation timing. The camera system supports socket-based attachment to skeletal bones, such as head, spine, or pelvis sockets, so first-person and actor-following views can inherit smooth character motion during jumping, falling, and landing. External character meshes are connected to the animation library through IK Rig and IK Retargeter assets, enabling different characters to share a consistent motion-control interface during counterfactual replay.

\begin{figure}[h]
  \centering
  \includegraphics[width=\linewidth]{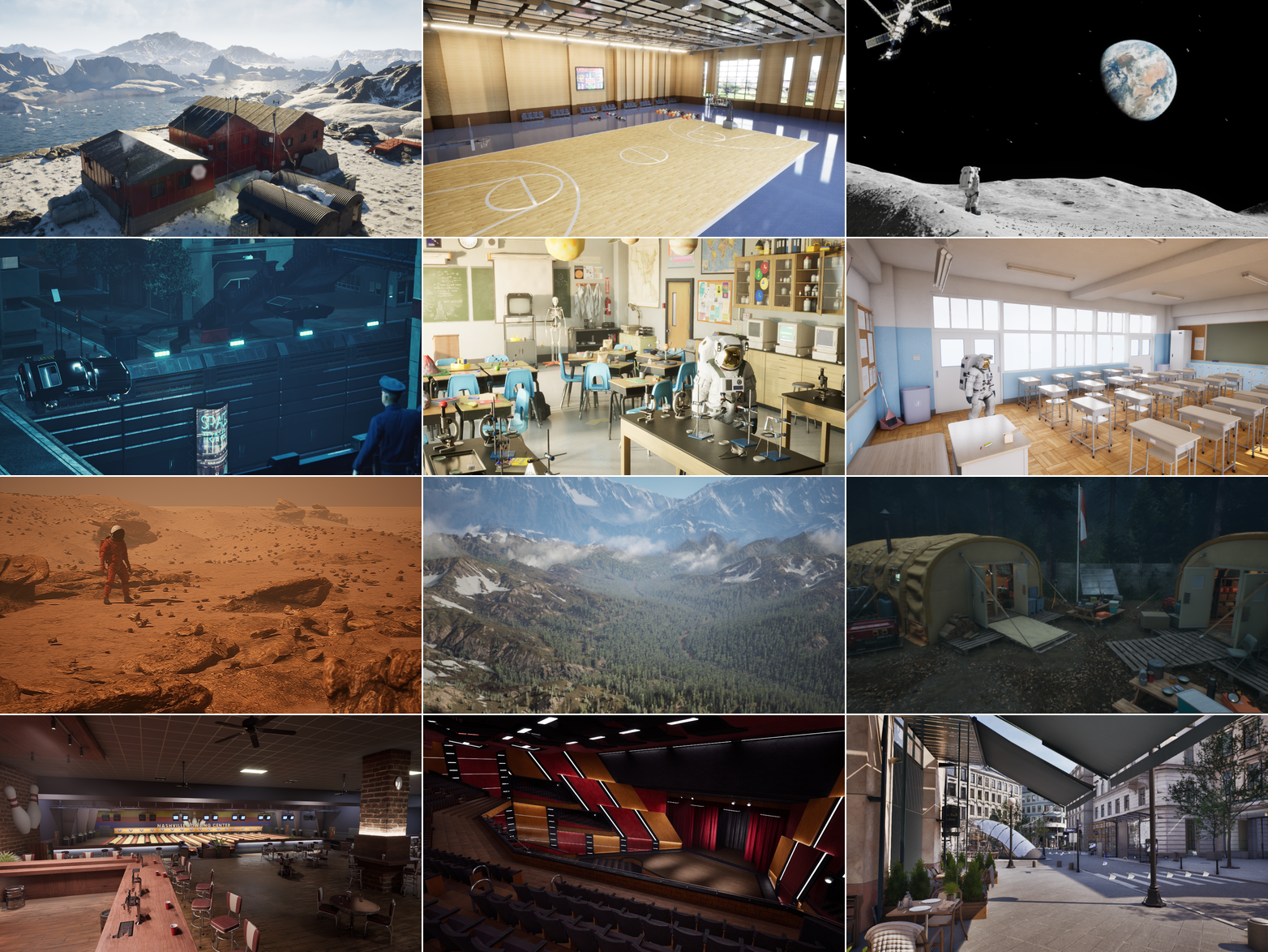}
  \caption{The 12 curated UE5 scene assets used in PhysEditWorld. The asset library covers indoor, outdoor, planetary, urban, sports, and object-interaction environments to support diverse gravity-sensitive motion patterns.}
  \label{fig:supp-scene-assets}
\end{figure}

\begin{figure}[t]
  \centering
  \includegraphics[width=0.85\linewidth]{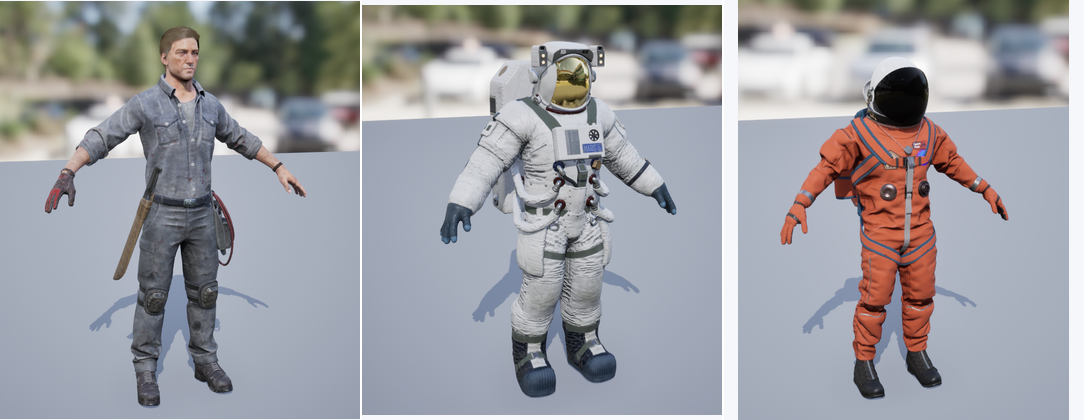}
  \caption{Character assets used in the current PhysEditWorld pipeline. Characters are configured with compatible controllers, animation retargeting, and low-gravity animation tuning to support reproducible replay under edited gravity settings.}
  \label{fig:supp-character-assets}
\end{figure}

PhysEditWorld is built from a manually curated UE5 asset library designed to cover diverse gravity-sensitive interaction scenarios. The current asset collection contains 12 representative scene environments, including polar research bases, indoor sports courts, lunar and Martian surfaces, cyberpunk city blocks, classrooms, laboratories, forest valleys, research camps, bowling alleys, theaters, and modern urban streets. These assets span indoor, outdoor, planetary, urban, sports, and object-interaction settings, allowing the dataset to expose different forms of gravity-dependent behavior such as free fall, jump arcs, landing timing, object displacement, and contact response. Artists manually inspect each scene for visual fidelity, collision quality, interaction suitability, and physical stability before it is included in the replay pipeline.

In addition to scene assets, PhysEditWorld includes character assets with compatible controller, animation, and retargeting setups. The character system uses a UE5 animation stack based on motion matching, pose search, chooser-driven animation selection, blend stacks, orientation warping, and animation warping. To support low-gravity scenarios, we implement a low-gravity animation tuner that adjusts airborne-state detection and animation playback rate according to the gravity multiplier. This prevents low-gravity clips from using visually implausible Earth-gravity animation timing. External character meshes are connected to the shared animation library through IK Rig and IK Retargeter assets, enabling different characters to share a consistent motion-control interface during counterfactual replay.

\subsection{Character-Centric Multi-Camera Capture}

The capture system uses a character-centric multi-camera rig rather than placing independent camera actors in the scene. Multiple \texttt{UCameraComponent}s are embedded inside the character blueprint and attached through skeletal sockets or bones, spring arms, and camera components. This topology provides a shared character-centered reference frame for all camera views, so first-person, third-person, side, front, back, and oblique views remain temporally aligned during walking, turning, jumping, falling, and landing. In the current setup, the camera array includes \texttt{FP}, \texttt{BK}, \texttt{FL}, \texttt{FW}, \texttt{FR}, \texttt{LF}, \texttt{RT}, and \texttt{TP} views. Rendering all views from the same replay instance avoids the synchronization drift that would arise from repeated simulation or separately placed cameras.

\begin{figure}[h]
  \centering
  \includegraphics[width=\linewidth]{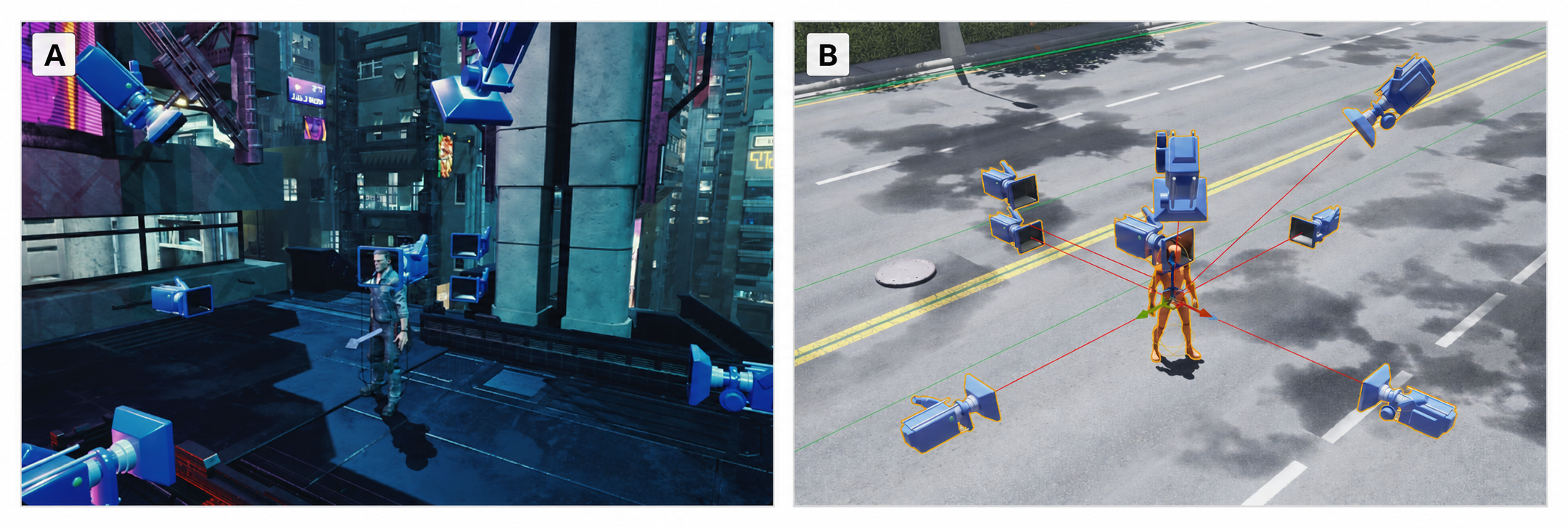}
  \caption{Character-centric multi-camera rig used for synchronized capture. (A) Cameras are embedded around the actor inside a UE5 scene. (B) The rig defines multiple named views around the same character reference frame, including first-person, third-person, side, front, and back views.}
  \label{fig:supp_multicamera_rig_re.png}
\end{figure}

Socket-based attachment also supports stable close-range observation. Cameras can be mounted to head, spine, pelvis, or other skeletal sockets, while spring arms decouple the choice of skeletal anchor from the camera offset and near-body framing. During interactive preview, spring arms can provide collision handling and visual smoothing. During dataset replay and rendering, camera lag is disabled to prioritize deterministic frame alignment and reproducible multi-view sampling. This design is especially important for gravity-edited rollouts, where jump and fall timing are the measurement target rather than merely cinematic motion.

\begin{figure}[h]
  \centering
  \includegraphics[width=\linewidth]{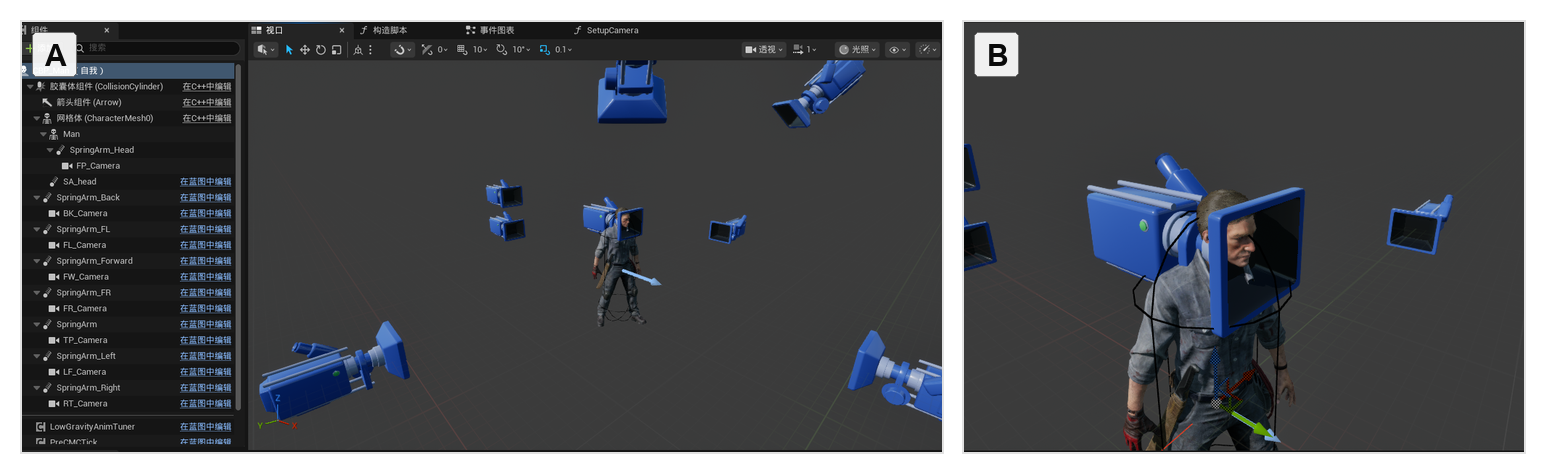}
  \caption{Socket and spring-arm based camera setup. (A) The character blueprint contains multiple spring-arm and camera components attached around the skeletal mesh. (B) Close-range cameras are mounted through skeletal sockets and spring arms, allowing first-person or near-body views to inherit character motion while preserving controllable camera offsets.}
  \label{fig:supp-camera-socket-setup}
\end{figure}

\subsection{Render Graph Outputs and Metadata}

\begin{figure}[h]
  \centering
  \includegraphics[width=\linewidth]{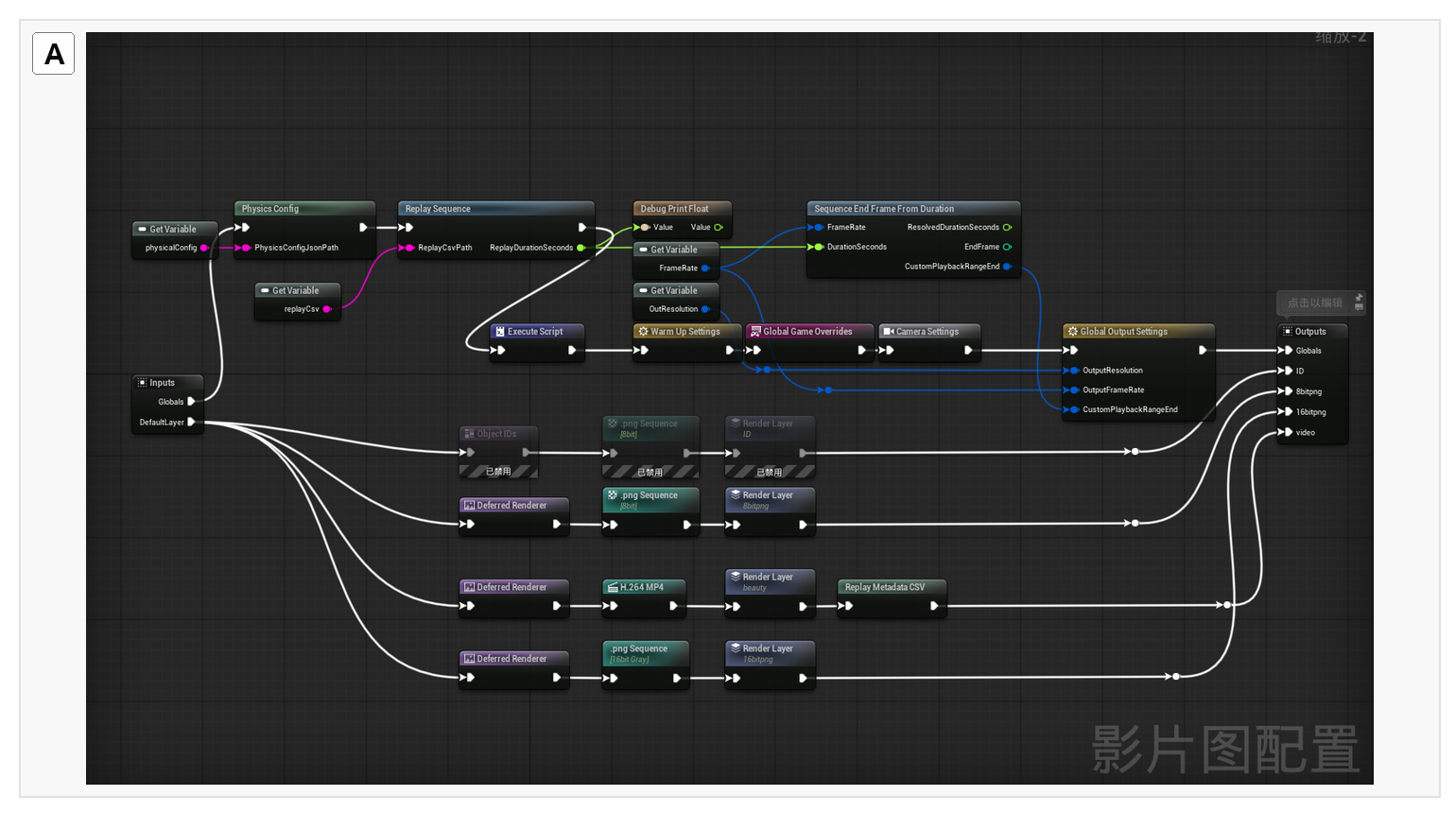}
  \caption{Movie Render Graph configuration for synchronized multimodal export. The graph combines replay sequence loading, warm-up, global game and camera settings, output settings, multi-branch visual exports, H.264 preview video, high-precision depth output, and replay metadata CSV export.}
  \label{fig:supp-render-export-graph}
\end{figure}

The UE5 rendering job is configured as a multi-branch Movie Render Graph. A shared global settings chain controls warm-up, game overrides, camera settings, and output settings. The current configuration uses a warm-up interval before formal capture so that animation state, temporal antialiasing, post-processing, and motion blur reach a stable state before frames are written. Captures are exported at $1280\times720$ resolution and 30 FPS.

The render graph produces several synchronized output branches from the same replay. An object-ID or segmentation branch exports per-frame ID supervision. An 8-bit image branch exports standard image sequences and auxiliary render passes such as motion vectors and world normals. A video branch exports H.264 MP4 previews with audio for manual inspection and semantic quality control. A 16-bit PNG branch exports high-precision depth-style supervision. In parallel, replay metadata is written as frame-level and time-level CSV files, together with a camera-name mapping JSON file. These metadata records include replay input events, player and camera transforms, physical state, camera identity, and capture termination signals.

\section{UE Editor Plugin and Data Generation Workflow}

PhysEditWorld is generated through a UE5 Editor plugin that manages the data-generation workflow from scene preparation to large-scale synthetic data production. The plugin is designed to operate directly on artist-authored UE5 levels with minimal intrusion into existing game content. Instead of rebuilding scenes in a separate simulator, it augments prepared levels with the runtime contexts, replay components, camera bindings, and batch-generation interfaces required for reproducible replay-and-rendering.

A key design goal of the plugin is to make scene preparation lightweight. As illustrated in Figure ??, each artist-prepared level can be converted into a production-ready data-generation scene through four editor operations: selecting the required EnhancedInputContext, specifying the sequence and camera bindings, running the FactoryManager initialization, and saving the produce-ready level. Since most updated character controller implementations already expose Enhanced Input support, this design remains extensible across different controllers and input schemes.

After the level-specific configuration is completed, the FactoryManager initializes the scene with a single command. It automatically registers the data-collection context, attaches replay components, inserts the required capture and rendering modules, and validates the camera and input bindings. This step injects only the minimal components needed for data generation, preserving the original authored scene and gameplay logic as much as possible. Once initialization finishes, the level becomes ready for reproducible replay and rendering.

The plugin then supports two ways of constructing interaction sequences. The first is Play-in-Editor interaction capture, where users directly play the prepared level and record semantic input-action traces, including movement axes, jump commands, camera deltas, and button states. The second is procedural sequence generation, where PCG-based scripts generate scalable interaction sequences for broader coverage. We also combined the Navmesh Agent in our pipline. to Both sources are converted into replayable interaction sequences that can be executed under controlled physical configurations.

For large-scale production, the DataFactory pipeline exposes Python interfaces and YAML configuration files for composing replay, rendering, cleaning, merging, and recovery stages. The batch system supports checkpointing and resume functionality, allowing interrupted jobs to continue from the last valid state rather than restarting from scratch. This makes the same editor-plugin workflow usable both for interactive debugging inside UE5 and for large-scale offline dataset generation.
\begin{figure}[H]
    \centering
    \includegraphics[width=1\linewidth]{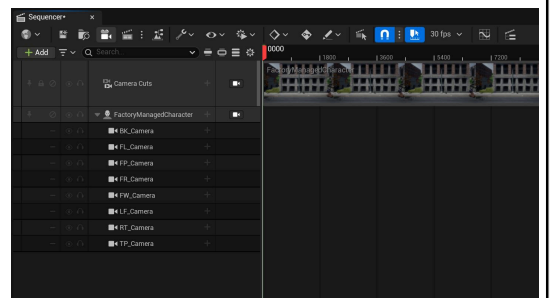}
    \caption{UE Editor Plugin and Data Generation Workflow}
    \label{fig:placeholder}
\end{figure}

\section{Dataset Sample Format and Metadata Schema}
\label{app:dataset_sample}

Each data sample in our dataset corresponds to one complete recording-rendering job under a fixed scene, action sequence, camera setup, and physical configuration. We use an 8-digit string as the sample identifier, e.g., \texttt{00000000}. A sample is self-contained and consists of synchronized multi-view RGB videos, frame-level metadata, event-level metadata, and a physical-configuration descriptor. The dataset root contains a global index file, \texttt{Details.json}, which stores the list of samples and the relative paths to their associated files.

The cleaned dataset follows the directory layout below:
\begin{verbatim}
<dataset_root>/
  Video/<sample_id>/<camera_name>.mp4
  Meta/<sample_id>_meta_frame.csv
  Meta/<sample_id>_meta_time.csv
  PhysicalConfig/<physical_config_name>.json
  Details.json
\end{verbatim}

For auxiliary modalities, such as depth, surface normals, or object masks, the optional frame-wise outputs are stored as:
\begin{verbatim}
Aux/<aux_type>/<sample_id>/<camera_name>/<frame_id>.png
\end{verbatim}

Table~\ref{tab:sample_schema} summarizes the main fields used to describe each sample.

\begin{table}[h]
\centering
\caption{Main fields of a dataset sample.}
\label{tab:sample_schema}
\begin{tabular}{ll}
\toprule
Field & Description \\
\midrule
\texttt{sample\_id} & Unique 8-digit identifier of the sample. \\
\texttt{scene\_name} & Name of the simulated scene or environment. \\
\texttt{action\_sequence\_name} & Identifier of the replayed action or motion sequence. \\
\texttt{physical\_config\_name} & Name of the physical condition used for simulation. \\
\texttt{physical\_config\_file} & Relative path to the physical-configuration JSON file. \\
\texttt{frame\_meta\_file} & Relative path to frame-level metadata. \\
\texttt{time\_meta\_file} & Relative path to event-level temporal metadata. \\
\texttt{camera\_names} & List of camera names available in the sample. \\
\texttt{cameras} & Per-camera video paths and attributes. \\
\texttt{aux\_types} & Optional auxiliary modalities associated with the sample. \\
\bottomrule
\end{tabular}
\end{table}

A representative sample entry is shown below. All paths are relative to the dataset root.

\begin{verbatim}
{
  "sample_id": "00000000",
  "scene_name": "Map_MarsRover",
  "action_sequence_name": "MarsSequence_0001",
  "physical_config_name": "G_0.05",
  "physical_config_file": "PhysicalConfig/G_0.05.json",
  "frame_meta_file": "Meta/00000000_meta_frame.csv",
  "time_meta_file": "Meta/00000000_meta_time.csv",
  "camera_names": ["BK_Camera", "FL_Camera", "FP_Camera", ... ],
  "aux_types": ["Depth", "Normals"],
  "cameras": [
    {
      "camera_name": "BK_Camera",
      "rgb_file": "Video/00000000/BK_Camera.mp4",
      "aux": {
        "Depth": "Aux/Depth/00000000/BK_Camera",
        "Normals": "Aux/Normals/00000000/BK_Camera",
        "ObjectMask": "Aux/ObjectMask/00000000/BK_Camera"
      },
      "resolution": {"width": 1280, "height": 720},
      "fps": 30
    },
    ...
  ]
}
\end{verbatim}

The two metadata files provide complementary temporal annotations. The frame-level file records information indexed by rendered frame, which is used to align visual observations with simulation states. The event-level file records simulation and capture events over time, such as the start and end of recording or replay. The physical-configuration file stores the controlled simulation parameters, enabling downstream methods to condition learning or evaluation on explicit physical settings.

\section{Example usage.}
The dataset generation pipeline is driven by a YAML configuration file and is launched through the repository-level entry point \texttt{Scripts/cli.py}. The configuration specifies the simulated scenes, replay trajectories, physical-configuration files, Movie Render Graph preset, output resolution, frame rate, and the enabled pipeline stages.

Before running the full pipeline, we first validate the configuration and inspect the planned jobs:
\begin{verbatim}
uv run --with pyyaml python Scripts/cli.py \
  --config Scripts/configs/base.yaml \
  --dry-run
\end{verbatim}

After verifying the generated plan, we run the complete rendering and cleaning pipeline:
\begin{verbatim}
uv run --with pyyaml python Scripts/cli.py \
  --config Scripts/configs/base.yaml
\end{verbatim}

A typical configuration contains the following fields:
\begin{verbatim}
base_path: <project_root>
unreal_exe: <path_to_UnrealEditor>
dataset_path: <raw_dataset_root>
output_path: <cleaned_dataset_root>
input_sequence: <replay_csv_root>
physical_config: <physical_config_root>

scenes:
  - name: <scene_name>
    level_path: <unreal_level_asset_path>
    level_sequence_path: <unreal_level_sequence_asset_path>

job:
  selection: auto
  mrg_path: <movie_render_graph_asset_path>
  params:
    fps: 30
    resolution: [1280, 720]
    mp4: true

pipeline:
  render:
    enable: true
    auto_start: true
  clean:
    enable: true
    input_root: <stage1_render_output_root>
    output_root: <cleaned_dataset_root>
    physical_config_root: <physical_config_root>
    append: false
    workers: 8
    overwrite: false
    dataset_name: <dataset_name>
\end{verbatim}

The command above expands the configuration into rendering jobs over the Cartesian product of scenes, replay trajectories, and physical configurations. It then launches Unreal Engine to execute the rendering jobs and, after rendering, converts the stage-1 outputs into the canonical dataset layout described in Appendix~\ref{app:dataset_sample}.

When the stage-1 rendering outputs have already been generated, the cleaning step can also be executed independently:
\begin{verbatim}
uv run python Scripts/utools/clean_mrq_graph_dataset.py \
  <stage1_render_output_root> \
  --output-root <cleaned_dataset_root> \
  --physical-config-root <physical_config_root> \
  --workers 8 \
  --dataset-name <dataset_name>
\end{verbatim}

This cleaning-only command scans the stage-1 output directories, identifies valid rendering jobs by the presence of \texttt{camera\_name\_map.json} and camera videos, copies the RGB videos and metadata into the canonical layout, and writes the global index file \texttt{Details.json}.

\section{More Asset}
A small set of data is showed in the supplementary material with a html for the limitation of size.

\begin{figure}[t]
    \centering
    \includegraphics[width=1\linewidth]{figures/Sup_More0.pdf}
    \caption{More PhysicsEdit-Data}
    \label{fig:placeholder}
\end{figure}
\begin{figure}[t]
    \centering
    \includegraphics[width=1\linewidth]{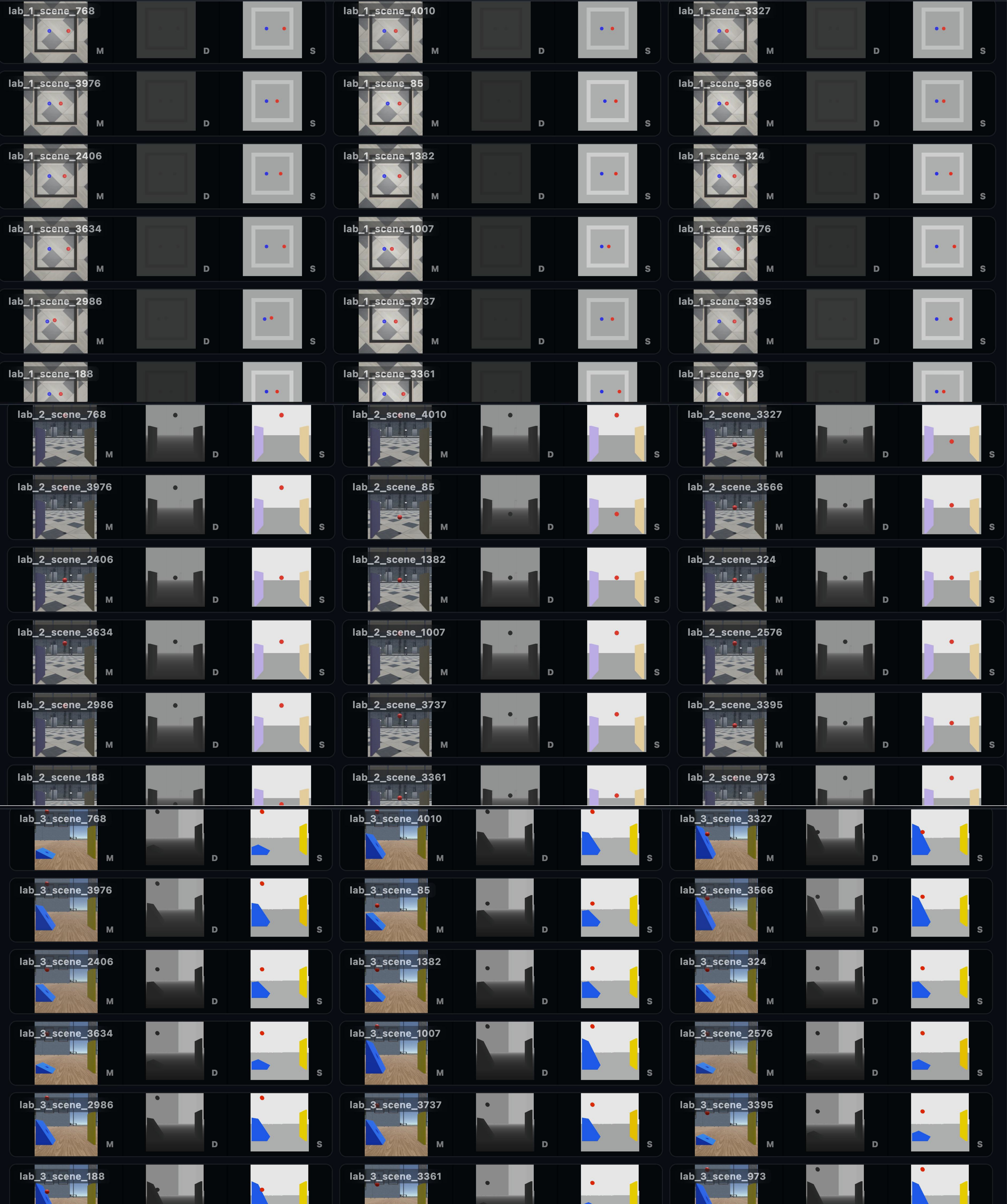}
    \caption{More PhysicsEdit-Data with other physics config like friction, bounce coefficient}
    \label{fig:placeholder}
\end{figure}
\begin{figure}[t]
    \centering
    \includegraphics[width=1\linewidth]{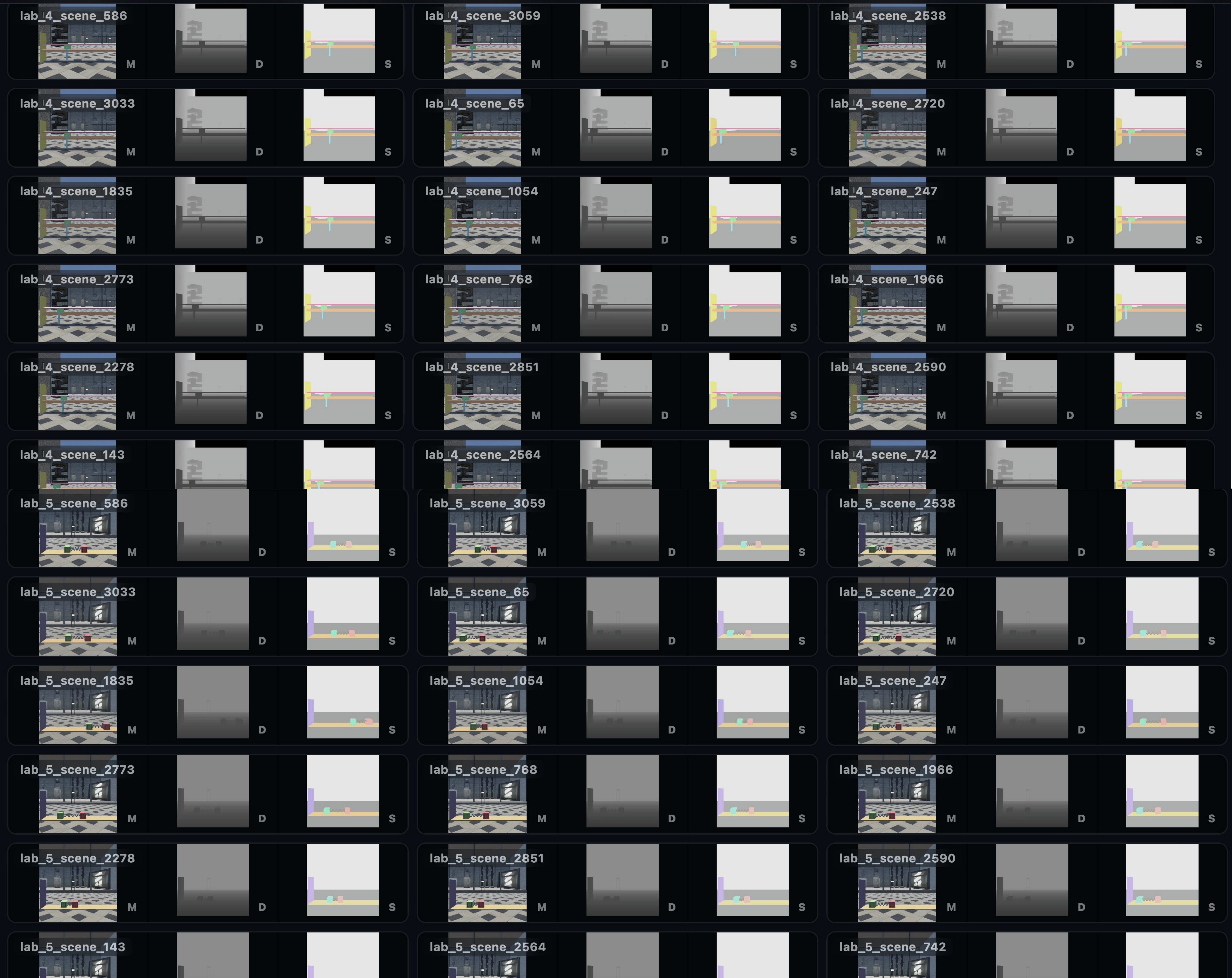}
    \caption{More PhysicsEdit-Data with other physics config like friction. bounce coefficient}
    \label{fig:placeholder}
\end{figure}

\section{Release Plan}
We plan to publicly release the PhysEditWorld dataset, including synchronized RGB videos, depth maps, normal maps, gravity annotations, camera trajectories, action traces, and evaluation scripts used in this work. We have provided a small subset in out supplementary material for the size limitation.

We also plan to release the UE5-based replay-and-rendering pipeline \textbf{as a plugin}, together with data generation configurations and example replay assets, to facilitate reproducibility and future research on physics-editable world modeling.Due to potential licensing restrictions associated with certain third-party UE5 assets, some raw scene assets or commercial content may not be redistributed directly. In such cases, we will provide replacement assets, asset lists, or reconstruction instructions whenever possible.

\clearpage

\end{document}